\documentclass[11pt]{article}

\usepackage[]{acl}

\usepackage{times}
\usepackage{latexsym}
\usepackage{CJKutf8}
\usepackage[T1]{fontenc}

\usepackage[utf8]{inputenc}

\usepackage{microtype}

\usepackage{inconsolata}

\usepackage{graphicx}
\usepackage{multirow}
\usepackage{multicol}
\usepackage{booktabs}
\usepackage{amsmath}
\usepackage[table]{xcolor}
\definecolor{lightgreen}{RGB}{220,245,220}
\definecolor{lightred}{RGB}{255,220,220}
\definecolor{lightgrey}{gray}{0.9}
\definecolor{midgreen}{RGB}{170,230,170}
\definecolor{midred}{RGB}{255,170,170}
\usepackage{hhline}
\newcommand{\benchname}{\textcolor{black}{CSM-MTBench}}

\newcommand{\zky}[1]{\textcolor{black}{#1}}
\usepackage[normalem]{ulem}

\title{Benchmarking Machine Translation on Chinese Social Media Texts}

\author{
 \textbf{Kaiyan Zhao\textsuperscript{1}},
 \textbf{Zheyong Xie\textsuperscript{2}},
 \textbf{Zhongtao Miao\textsuperscript{1}}, 
 \textbf{Xinze Lyu\textsuperscript{2}},
 \textbf{Yao Hu\textsuperscript{2}},
 \textbf{Shaosheng Cao\textsuperscript{2,3}\thanks{Corresponding author.}}
\\
 \textsuperscript{1}The University of Tokyo,
 \textsuperscript{2}Xiaohongshu Inc.,
 \textsuperscript{3}Tsinghua University
\\
\normalfont{\fontsize{11pt}{12pt}\selectfont {\fontfamily{qcr}\selectfont caoshaosheng@xiaohongshu.com}} \\
}

\begin{document}
\begin{CJK*}{UTF8}{gbsn}

\maketitle
\begin{abstract}
The prevalence of rapidly evolving slang, neologisms, and highly stylized expressions in informal user-generated text, particularly on Chinese social media, poses significant challenges for Machine Translation~(MT) benchmarking. Specifically, we identify two primary obstacles:
(1) data scarcity, as high-quality parallel data requires bilingual annotators familiar with platform-specific slang, and stylistic cues; and
(2) metric limitations, where traditional evaluators like COMET often fail to capture stylistic fidelity and non-standard expressions.
To bridge these gaps,
we first introduce \benchname, an expert-curated benchmark covering five Chinese$\to$foreign language directions \zky{and comprising 10,915 total instances}, organized into two subsets: \textbf{Fun Posts}, featuring context-rich content with frequent slang and neologisms, and \textbf{Social Snippets}, focusing on concise, emotion- and style-driven expressions.
Next, we propose tailored evaluation approaches for each subset,
\zky{including slang translation success rate and style-preservation metrics combining embeddings and LLM-as-a-judge evaluation.}
Experiments on over 20 models reveal substantial variation in how current MT systems handle semantic fidelity, slang translation, and stylistic nuances. 
\zky{We further look into model behaviors through prompt sensitivity analysis and qualitative case studies, highlighting the challenges of translating these informal expressions.}
\benchname~thus serves as a rigorous testbed for advancing MT systems capable of mastering real-world Chinese social media texts.\footnote{ We release the dataset at \url{https://github.com/KYuuto1006/CSM-MTBench}.}
\end{abstract}

\section{Introduction}
Machine Translation~(MT) has long been one of the central tasks in Natural Language Processing~(NLP), as it enables communication and knowledge exchange across languages~\citep{neuralmt, newman-griffis-etal-2021-translational}. Despite recent advances with Large Language Models (LLMs)~\citep{wang-etal-2023-document-level-MT, zhu-etal-2024-multilingual, xu2024alma, wu-etal-2024-wordalignmentmt}, translating informal user-generated content on social media remains challenging~\citep{qian-etal-2024-evaluatingweibo5538, guo2025redefiningmachinetranslationsocial}.

\begin{figure*}[ht]
    \centering  \includegraphics[width=1.0\textwidth]{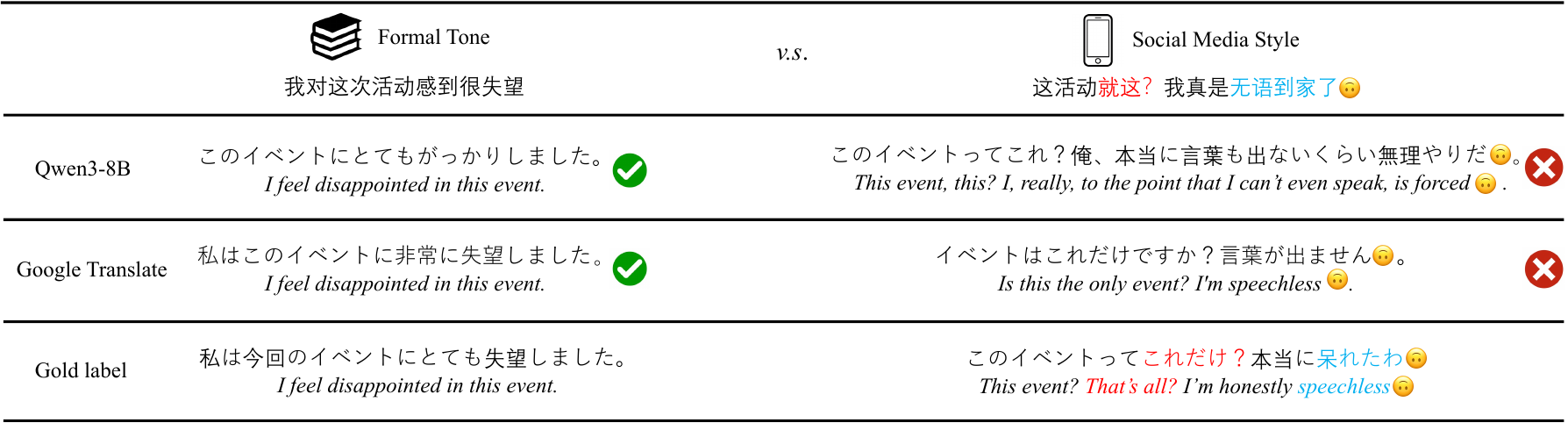}
    \caption{MT into Japanese using Qwen3-8B and Google Translation for two Chinese source sentences expressing the same meaning (``I feel disappointed in this event.'') in formal tone versus social media style. English references are shown in \textit{italics}. Social-media-specific tone is highlighted in \textcolor{red}{red}, and slang is highlighted in \textcolor{cyan}{blue}.}
    \label{fig:1}
\end{figure*}

Unlike the formal texts (e.g., news or Wikipedia) commonly used in existing MT benchmarks~\citep{guzman-etal-2019-flores, goyal-etal-2022-flores101}, social media texts exhibit distinct linguistic characteristics, including rapidly evolving slang, informal expressions, and highly stylized language~\citep{al-amer-etal-2025-comparative}.
This phenomenon is particularly prominent on Chinese social media platforms, where users frequently create neologisms, abbreviations, and novel expressions through creative recombinations of existing characters, many of which evolve rapidly over time and are rarely captured by static MT benchmarks~\citep{miao2026neoamtneologismawareagenticmachine, zhao2026cneobenchdiagnosinglargelanguage}.
Figure~\ref{fig:1} illustrates an example where MT models translate two semantically equivalent Chinese sentences of distinct styles, formal versus social media, into Japanese.
While the models successfully handle the formal sentence, they fail to preserve the slang and the tone specific to social media, highlighting the challenges of translating these informal and creatively recombined\footnote{For example, the Chinese phrase ``无语到家了'' (really speechless) in Figure~\ref{fig:1} creatively combines ``无语'' (speechless)'' with ``到家了'' (back to home), forming a new slang.} user-generated content.

To build a benchmark for Chinese social media MT, we identify two main challenges: data scarcity and evaluation limitations. Existing Chinese social media resources such as BOLT~\citep{song-etal-2014-collecting-bolt} and HADQAET~\citep{qian2023evaluationchineseenglishmachinetranslation} largely reflect earlier-stage social-media languages that can be easily handled by recent LLMs. Meanwhile, constructing high-quality annotations for up-to-date social media data requires bilingual annotators who are familiar with social media expressions in both languages, a skill set that is not widely available. In addition, traditional automatic metrics including BLEU~\citep{papineni-etal-2002-bleu} and chrF~\citep{popovic-2015-chrf} and COMET~\citep{rei-etal-2020-comet} are often insufficient for evaluating translations of informal user-generated content where meaning may be conveyed through special expressions.


To address these challenges, we introduce a new benchmark, \benchname, comprising 10,915 total translation samples for systematically evaluating MT performance on authentic Chinese social media texts. The up-to-date source data are collected from a real Chinese social media platform and manually translated into Spanish~(es), French~(fr), Japanese~(ja), Korean~(ko) and Russian~(ru) by expert human annotators, ensuring high-quality multilingual references for evaluation. 
Specifically, \benchname~consists of two complementary datasets: \textbf{Fun Posts} and \textbf{Social Snippets}. Fun Posts contain longer, content-rich \underline{user posts} describing events or experiences and often include slang and neologisms. In contrast, Social Snippets capture short, highly emotional \underline{comments} that exhibit distinctive tones and stylistic features. Together, these datasets reflect the diverse linguistic styles in Chinese social media.

To better evaluate the MT performance, we combine standard evaluation metrics such as XCOMET~\citep{guerreiro-etal-2024-xcomet} with targeted evaluation methods tailored to the distinct challenges of each dataset.
For Fun Posts, we focus on the accurate translation of neologisms and slang, leveraging a carefully curated slang dictionary and fuzzy matching to determine whether candidate translations appear in the model output. For Social Snippets, we evaluate the preservation of social-media-specific tone and style by combining style embeddings~\citep{patel-etal-2025-styleembedding}, emotion embeddings~\citep{poulaei-etal-2025-ynwa-emotionembedding}, and sentiment embeddings~\citep{tabularisai_2025-sentimentembed}, complemented by LLM-as-a-judge~\citep{gu2025surveyllmasajudge} method to assess tone and style consistency.

We evaluate over 20 different models on \benchname, including both MT-specific models and general-purpose LLMs. Our results reveal several notable trends. While LLMs achieve strong overall performance on Fun Posts according to standard metrics, their translations of neologisms and slang remain inconsistent. On Social Snippets, models often fail to preserve the tone and style of the source, particularly in emotionally charged content. 
These findings highlight the challenges of translating real-world Chinese social media and the need for benchmarks and evaluation methods that better capture these linguistic phenomena.
\zky{We further conduct prompt sensitivity experiments and qualitative analyses to better understand model behaviors, and find that even simple sentences become difficult for models to translate when additional emotional features, such as emojis or repeated exclamation marks are present.}


\section{Related Work}
\subsection{Machine Translation with LLMs}

Initial attempts of applying general-purpose LLMs to MT, including models such as GPT3~\citep{GPT3} and BLOOM~\citep{workshop2023bloom176bparameteropenaccessmultilingual}, demonstrate strong multilingual capabilities and promising zero-shot and few-shot performance of LLMs. Subsequent work shows that prompting plays a crucial role in unlocking LLMs’ translation ability~\citep{vilar-etal-2023-promptingpalm, zhang2023promptinglargelanguagemodel, ghazvininejad2023dictionarybasedphraselevelpromptinglarge}, while studies such as \citet{chitale-etal-2024-empiricalICL} and \citet{lee-etal-2025-exploring-ICL} further establish the effectiveness of in-context learning~(ICL) for MT.
The Aya family~\citep{2024aya101modelinstructionfinetuned, aryabumi2024aya23openweight, dang2024ayaexpansecombiningresearch} represents a line of LLMs optimized explicitly for multilingual abilities.

In parallel, recent research has explored reinforcement learning for MT specific models~\citep{CPO, wu-etal-2024-wordalignmentmt, guo2025redefiningmachinetranslationsocial}. 
The Tower series models~\citep{alves2024toweropenmultilinguallarge} are proposed to deal with multiple tasks present in translation workflows. GemmaX2~\citep{gemmax2cui2025multilingualmachinetranslationopen} incorporates data-mixing strategies during continual pretraining to enhance multilingual performance, while Hunyuan-MT~\citep{zheng2025hunyuanmttechnicalreport} employs carefully engineered pre-training and post-training pipelines tailored specifically for MT.

\subsection{MT on Social Media Texts}
\zky{Prior work has explored MT for informal text in several languages. The NUS SMS Corpus~\citep{Chen_2012_nus_sms} contains short message data, reflecting earlier chat-style informal texts. OpenSubtitles~\citep{lison-tiedemann-2016-opensubtitles2016} provides conversational parallel data derived from movie subtitles, which exhibit spoken language characteristics. \citet{michel-neubig-2018-mtnt} introduce MTNT, a parallel corpus of noisy social media text collected from Reddit, designed to evaluate English, French and Japanese MT performance. MMTC~\citep{mcnamee-duh-2022-multilingual_mmtc} is constructed from multilingual Twitter posts and focuses on translation of user-generated microblog content.}

\zky{Closest to our setting, several datasets have been proposed for Chinese social media MT. \citet{ling-etal-2013-microblogs} rely on automatic mining methods for microblogs data. BOLT~\citep{song-etal-2014-collecting-bolt} collects informal chat data from online messaging platforms. }
\citet{qian-etal-2024-evaluatingweibo5538} construct a Weibo\footnote{A mainstream Chinese social platform.}-based dataset, with evaluation restricted to the zh$\to$en direction. \citet{guo2025redefiningmachinetranslationsocial} propose Redtrans-Bench, which contains 2858 zh$\to$en SNS~(Social Networking Service) translation pairs. However, the dataset is limited to a single language pair, and evaluation is only performed using XCOMET~\citep{guerreiro-etal-2024-xcomet}, without consideration for SNS-specific features.


\zky{\benchname~differs from existing resources in four important aspects: (1) it is constructed from up-to-date Chinese social media data, capturing the rapidly evolving slang and stylistic conventions on modern platforms; (2) it covers multiple target languages beyond the commonly studied zh$\to$en direction; (3) a  more careful curation process with human annotations; and (4) targeted evaluation metrics for social-media texts besides traditional ones.}

\subsection{Domain-specific MT}
For Chinese-specific domains, recent research has explored MT challenges arising from culturally rich or linguistically unique content. Several works focus on Chinese idiom translation~\citep{idiomkb, fu-etal-2025-chengyu, yang2025evaluatingchineseidiom}, a domain where fixed expressions and metaphorical meanings pose difficulties for both neural MT systems and LLMs. WYWEB~\citep{zhou-etal-2023-wyweb} introduces a benchmark for classical ancient Chinese, covering tasks such as sentence classification and MT. Similarly, \citet{chen-etal-2025-benchmarking-llms} propose PoetMT for translating classical Chinese poetry into English.

For culturally related domains more broadly, \citet{yao-etal-2024-benchmarking-cultural-awareness} evaluate MT on culture-specific entities and concepts across languages. NEO-BENCH~\citep{zheng-etal-2024-neobench} examines model robustness to neologisms, with MT evaluation on en$\to$zh samples. NewTerm~\citep{lerner-yvon-2025-towardsscientificneo} focuses on scientific neologisms which covers en$\to$fr translation. 

\section{\benchname}
\label{benchmark}
In this section, we first describe the construction of \benchname, followed by detailed explanations of the two subsets: Fun Posts and Social Snippets, and their corresponding evaluation metrics.

\subsection{Dataset Construction}
\label{dataset_construction}
\begin{figure*}[ht]
    \centering  \includegraphics[width=0.99\textwidth]{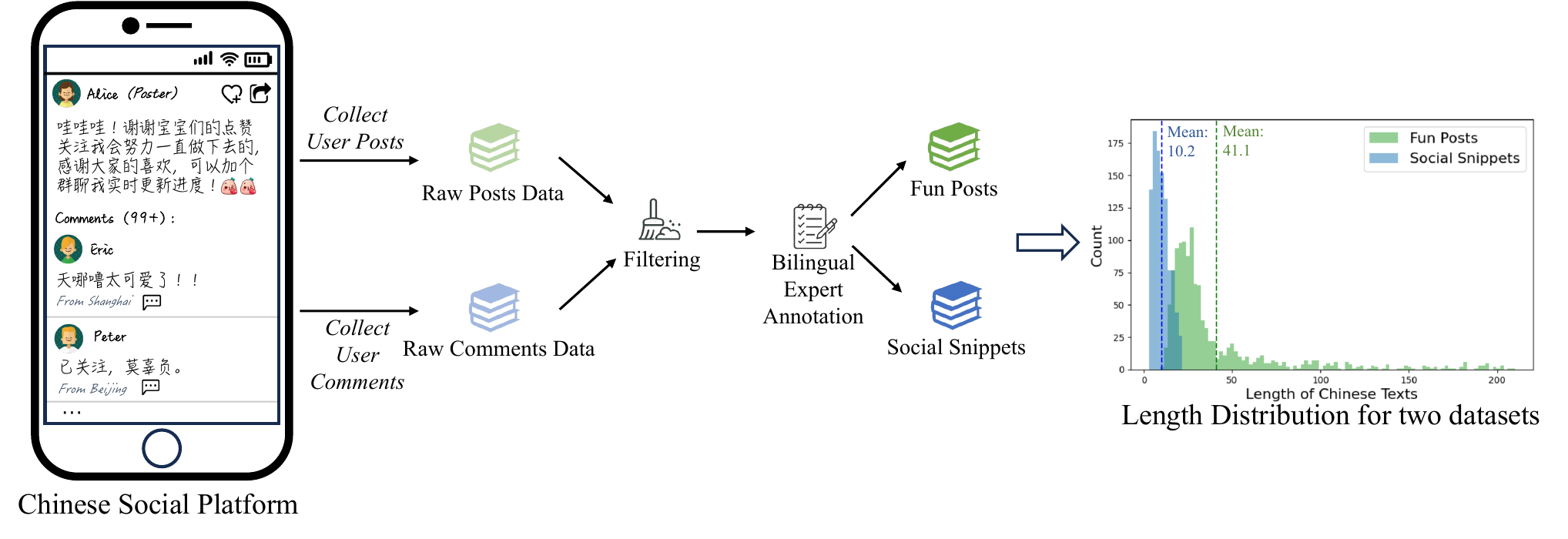}
    \caption{Construction of \benchname. We collect real-world data from a Chinese social platform and split them into Fun Posts and Social Snippets. All data undergo filtering and bilingual expert translation.}
    \label{fig:2}
\end{figure*}

\zky{The source data of \benchname~are collected from Xiaohongshu\footnote{Also known as RedNote, see \url{https://en.wikipedia.org/wiki/Xiaohongshu}}, a popular Chinese social media platform where users publish posts and interact through comments. We first collect a large pool of candidate data from January 2025 to July 2025, including over 100k posts and comments.} 

\zky{Unlike prior work that typically relies on random sampling through platform APIs, we adopt an engagement-aware data construction strategy to better capture representative social-media language. For posts, we select 1,500 posts as candidate source texts by considering exposure level, view counts, and the number of likes, aiming to prioritize widely viewed and interacted posts that reflect commonly encountered language patterns. For comments, based on engagement signals (e.g., likes, replies) and expressiveness of the text, we initially selected 5,000 candidate comments.}

\begin{figure}[t]
    \centering  \includegraphics[width=0.5\textwidth]{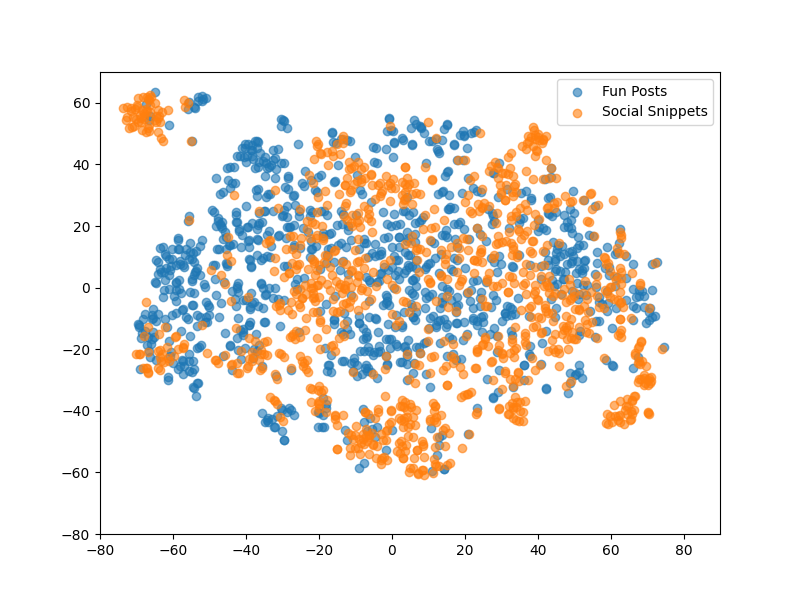}
    \caption{t-SNE illustration for two subsets.}
    \label{tsne}
\end{figure}

\zky{We then separate user posts and comments into two subsets: Fun Posts and Social Snippets (Figure~\ref{fig:2}; English translations of examples are in Appendix, Figure~\ref{fig:2 translation}). The candidate data undergo filtering before being translated into Spanish, French, Japanese, Korean, and Russian by bilingual experts. After filtering, 1,183 samples are retained in Fun Posts and 1,000 in Social Snippets. Dataset statistics are summarized in Table~\ref{statistics}, and Figure~\ref{tsne} visualizes the t-SNE figure for the two subsets using Qwen3-Embedding-8B. The two subsets exhibit partial separation with limited overlap, indicating noticeable distributional differences. Quantitatively, the cosine similarity between the subset centroids is approximately 0.44, and the normalized centroid distance is 0.24. These values suggest a meaningful semantic distribution shift rather than merely minor stylistic variation.
Additional details on the data processing pipeline are provided in Appendix~\ref{appendixA1} and \ref{annotation_discussion}.}

\begin{table}[t]
\centering
\resizebox{0.48\textwidth}{!}{
\begin{tabular}{lccc}
\hline
Subset           & Pair num. & Total num. & Total Samples        \\\hline
Fun Posts              & 1,183          & 5,915       & \multirow{2}{*}{10,915} \\
Social Snippets        & 1,000          & 5,000       &     \\ \hline                 
\end{tabular}}
\caption{Statistics for \benchname. Pair nums stands for the number of samples per source-target pair. }
\label{statistics}
\end{table}

\subsection{Descriptive Side: Fun Posts}

As shown in the right part of Figure~\ref{fig:2}, Fun Posts consist of longer and more coherent user posts that resemble short narratives or mini-blogs, with an average length of 41.1 Chinese characters. These sentences are generally complete and information-rich, often containing neologisms and slang (see Appendix, Figure~\ref{fig:example}). Fun Posts thus represent the descriptive side of social media, where successful translation requires preserving semantic nuances.


To quantify the preservation of these expressions, we introduce the Slang Success Rate (SSR), following~\citet{semenov-etal-2025-findings-WMT25}. Given the $i_{th}$ annotated source–target pair $(S_i, T_i)$, we first prompt GPT-5 to identify slang or neologistic expressions $s$ in the Chinese source sentence and locate their corresponding translations $t$ in the gold target sentence. This produces a set of slang–translation pairs:
\begin{equation}
    \mathcal{P}_i = [(s_1, t_1),..., (s_k, t_k)],
\end{equation}
where $k$ denotes the number of slang items in the $i_{th}$ source-target pair.

Based on these slang-translation pairs, we further prompt GPT-5 to augment each pair into a candidate dictionary, where multiple plausible target-language expressions are generated under the given context. 
For each slang item $s_j$, GPT-5 generates a list of valid translation candidates, denoted as $\mathcal{C}_j$:
\begin{equation}
    \mathcal{C}_j = [candi_j^1,..., candi_j^m],
\end{equation}
where $m$ is the number of generated candidate expressions. 
We then define the full set of acceptable translation candidates for $s_j$ as $\mathcal{C}_j^+=[t_j, candi_j^1,..., candi_j^m]$.
The prompt employed to identify slang-translation pairs $\mathcal{P}$ and generate these candidates is presented in Figure~\ref{fig:candidate}, Appendix. Subsequently, human annotators review the generated candidates to filter out invalid entries. The final data structure for Fun Posts is illustrated in the upper part of Figure~\ref{fig:example}.


After obtaining the model translation $\hat{T}_i$, we apply fuzzy matching\footnote{\url{https://github.com/rapidfuzz/RapidFuzz}}
to determine whether any expression in $\mathcal{C}_j^+$ appears in the output. The score for slang item $s_j$ is defined as:
\begin{equation}
\resizebox{0.4\textwidth}{!}{$
    \mathrm{Score}_{ij} = \left\{
    \begin{aligned}
    & 1, \mathrm{if} \ FM(\hat{T}_i, \mathcal{C}_j^+) \geq threshold   \\
    & 0, \ \mathrm{otherwise} \\
    \end{aligned}
    \right.,
$}
\end{equation}
where $FM(\cdot)$ denotes the fuzzy matching function. Based on these scores, the Slang Success Rate is defined as:
\begin{equation}
    SSR = \frac{1}{N}\displaystyle
    \sum_{i=1}^{N} \frac{1}{k}\sum_{j=1}^{k} \mathrm{Score}_{ij},     
\end{equation}

where $N$ denotes the number of source–target pairs containing slang or neologisms.

In total, 619 of the 1,183 Fun Posts samples contain detected slang or neologisms, and each slang term has 6.6 candidate translations on average.



\subsection{Reactive Side: Social Snippets}

Social Snippets, by contrast, capture the reactive and highly informal side of social media communication. These entries are typically short, sometimes only a few characters, and often rely on tone markers and stylized expressions that convey meaning beyond the literal words. Much of the content is emotion- or attitude-driven, expressing reactions such as laughter, surprise, or frustration rather than conveying factual information (see Appendix, Figure~\ref{fig:example}). As a result, Social Snippets provide a challenging testbed for MT systems, where standard metrics such as XCOMET~\citep{guerreiro-etal-2024-xcomet} may not fully capture the informal and nuance-rich characteristics of the text.


Unlike Fun Posts, the lexical simplicity of Social Snippets makes it difficult to isolate interpretable slang or neologisms for evaluation (see Appendix~\ref{difficulty}). To address this issue, we adopt an embedding-based evaluation that measures whether translations preserve stylistic and emotional properties of the source. Specifically, we combine three types of embeddings: style embeddings~\citep{patel-etal-2025-styleembedding}, emotion embeddings~\citep{poulaei-etal-2025-ynwa-emotionembedding}, and sentiment embeddings~\citep{tabularisai_2025-sentimentembed}.


Given a source sentence $S$ and model translation $\hat{T}$, we encode them using the embedding models and compute the cosine similarity between them:
\begin{equation}
    e_S, e_{\hat{T}} = \mathrm{EmbeddingModel}(S, \hat{T}),
\end{equation}
\vspace{-5mm}
\begin{equation}
    sim = \mathrm{CosineSim}(e_S, e_{\hat{T}}). 
\end{equation}
The final Embedding Similarity (ES) score is obtained by averaging the cosine similarities across the three embedding types, providing a holistic measure of how well the translation preserves the source’s style, emotion, and sentiment.


To further leverage our annotated translations, we also adopt a GEMBA-based evaluation method~\citep{kocmi-federmann-2023-large-gemba}. Specifically, we use the GEMBA-stars prompting framework (see Appendix Figure~\ref{fig:gemba}), where an LLM judges whether the tone and style of the source sentence are preserved in the translation.

\section{Experiments}
\begin{table*}[t]
\centering
\resizebox{0.99\textwidth}{!}{
\begin{tabular}{lcccccccccccc}
\toprule
\multicolumn{1}{c|}{\multirow{2}{*}{Models}} & \multicolumn{2}{c|}{$zh \to es$}          & \multicolumn{2}{c|}{$zh \to fr$}          & \multicolumn{2}{c|}{$zh \to ja$}          & \multicolumn{2}{c|}{$zh \to ko$}          & \multicolumn{2}{c|}{$zh \to ru$}          & \multicolumn{2}{c}{avg.} \\ \cmidrule(lr){2-3} \cmidrule(lr){4-5} \cmidrule(lr){6-7} \cmidrule(lr){8-9} \cmidrule(lr){10-11} \cmidrule(lr){12-13}
\multicolumn{1}{c|}{}                        & XCOMET & \multicolumn{1}{c|}{SSR}   & XCOMET & \multicolumn{1}{c|}{SSR}   & XCOMET & \multicolumn{1}{c|}{SSR}   & XCOMET & \multicolumn{1}{c|}{SSR}   & XCOMET & \multicolumn{1}{c|}{SSR}   & XCOMET      & SSR        \\ \midrule
\multicolumn{13}{c}{\textit{Closed-source LLM APIs}}                                                                                                                                                                                                                               \\ \midrule
\multicolumn{1}{l|}{GPT-4o}                   & \textbf{86.38}  & \multicolumn{1}{c|}{69.14} & \textbf{83.06}  & \multicolumn{1}{c|}{68.01} & \textbf{88.21}  & \multicolumn{1}{c|}{63.17} & 85.36  & \multicolumn{1}{c|}{60.58} & \textbf{84.49}  & \multicolumn{1}{c|}{63.65} & \textbf{85.50}       & 64.91      \\
\multicolumn{1}{l|}{GPT-5}                    & \underline{86.34}  & \multicolumn{1}{c|}{\textbf{79.32}} & \underline{82.67}  & \multicolumn{1}{c|}{\textbf{81.09}} & 87.64  & \multicolumn{1}{c|}{\textbf{75.12}} & \underline{85.46}  & \multicolumn{1}{c|}{\textbf{76.58}} & \underline{84.17}  & \multicolumn{1}{c|}{\textbf{76.41}} & \underline{85.26}       & \textbf{77.70}      \\
\multicolumn{1}{l|}{Claude-Sonnet-4}         & 86.05  & \multicolumn{1}{c|}{\underline{74.15}} & 82.49  & \multicolumn{1}{c|}{\underline{73.34}} & \underline{87.81}  & \multicolumn{1}{c|}{\underline{66.72}} & \textbf{85.54}  & \multicolumn{1}{c|}{\underline{72.21}} & 84.07  & \multicolumn{1}{c|}{\underline{70.11}} & 85.19       & \underline{71.31}      \\ \midrule
\multicolumn{13}{c}{\textit{Translation-specialized Models}}                                                                                                                                                                                                                       \\ \midrule
\multicolumn{1}{l|}{NLLB-3.3B}               & 70.07  & \multicolumn{1}{c|}{27.95} & 64.50  & \multicolumn{1}{c|}{27.14} & 61.52  & \multicolumn{1}{c|}{18.26} & 64.28  & \multicolumn{1}{c|}{21.16} & 67.09  & \multicolumn{1}{c|}{23.10} & 65.49       & 23.52      \\
\multicolumn{1}{l|}{google-translate}        & 83.63  & \multicolumn{1}{c|}{\textbf{58.97}} & 79.87  & \multicolumn{1}{c|}{\textbf{55.57}} & 82.90  & \multicolumn{1}{c|}{49.43} & 81.55  & \multicolumn{1}{c|}{48.95} & \underline{81.72}  & \multicolumn{1}{c|}{\textbf{54.93}} & 81.93       & \textbf{53.57}      \\
\multicolumn{1}{l|}{GemmaX2-9B}                 & \underline{84.06}  & \multicolumn{1}{c|}{\underline{57.03}} & \textbf{80.18}  & \multicolumn{1}{c|}{\underline{55.41}} & \underline{84.94}  & \multicolumn{1}{c|}{\textbf{49.76}} & \underline{82.95}  & \multicolumn{1}{c|}{\underline{49.60}} & 81.45  & \multicolumn{1}{c|}{\underline{51.59}} & \underline{82.72}       & \underline{52.68}      \\
\multicolumn{1}{l|}{Hunyuan-MT-7B}           & \textbf{84.35}  & \multicolumn{1}{c|}{54.60} & \underline{80.17}  & \multicolumn{1}{c|}{52.18} & \textbf{86.31}  & \multicolumn{1}{c|}{\underline{49.60}} & \textbf{85.05}  & \multicolumn{1}{c|}{\textbf{51.70}} & \textbf{82.66}  & \multicolumn{1}{c|}{49.92} & \textbf{83.71}       & 51.60      \\ \midrule
\multicolumn{13}{c}{\textit{Open-source General-purpose LLMs}}                                                                                                                                                                                                                             \\ \midrule
\multicolumn{1}{l|}{DeepSeek-V3}             & \underline{84.33}  & \multicolumn{1}{c|}{\textbf{70.11}} & \underline{81.05}  & \multicolumn{1}{c|}{\textbf{70.92}} & \textbf{87.26}  & \multicolumn{1}{c|}{\textbf{67.04}} & \underline{84.34}  & \multicolumn{1}{c|}{\textbf{64.94}} & \textbf{83.50}  & \multicolumn{1}{c|}{\textbf{69.31}} & \textbf{84.10}       & \textbf{68.46}      \\
\multicolumn{1}{l|}{GPT-OSS-120B}            & 84.25  & \multicolumn{1}{c|}{65.11} & \underline{81.05}  & \multicolumn{1}{c|}{62.04} & 85.47  & \multicolumn{1}{c|}{58.64} & 83.80  & \multicolumn{1}{c|}{58.16} & 81.62  & \multicolumn{1}{c|}{56.22} & \underline{83.24}       & 60.03      \\
\multicolumn{1}{l|}{Aya-101}                 & 74.26  & \multicolumn{1}{c|}{29.89} & 69.45  & \multicolumn{1}{c|}{28.59} & 77.36  & \multicolumn{1}{c|}{30.86} & 72.89  & \multicolumn{1}{c|}{26.82} & 72.32  & \multicolumn{1}{c|}{29.08} & 73.26       & 29.05      \\
\multicolumn{1}{l|}{Aya-Expanse-8B}          & 82.31  & \multicolumn{1}{c|}{56.54} & 78.25  & \multicolumn{1}{c|}{54.28} & 83.42  & \multicolumn{1}{c|}{51.86} & 81.64  & \multicolumn{1}{c|}{48.47} & 78.70  & \multicolumn{1}{c|}{52.10} & 80.86       & 52.65      \\
\multicolumn{1}{l|}{Gemma3-4B}               & 79.72  & \multicolumn{1}{c|}{53.15} & 74.98  & \multicolumn{1}{c|}{48.79} & 80.61  & \multicolumn{1}{c|}{42.49} & 77.67  & \multicolumn{1}{c|}{44.43} & 76.83  & \multicolumn{1}{c|}{46.37} & 77.96       & 47.05      \\
\multicolumn{1}{l|}{Gemma3-12B}              & 83.21  & \multicolumn{1}{c|}{56.38} & 78.84  & \multicolumn{1}{c|}{57.35} & 83.88  & \multicolumn{1}{c|}{51.05} & 81.25  & \multicolumn{1}{c|}{54.60} & 81.13  & \multicolumn{1}{c|}{57.18} & 81.66       & 55.31      \\
\multicolumn{1}{l|}{Gemma3-27B}              & 84.49  & \multicolumn{1}{c|}{64.29} & 80.73  & \multicolumn{1}{c|}{60.90} & 85.65  & \multicolumn{1}{c|}{56.22} & 83.13  & \multicolumn{1}{c|}{55.09} & \underline{82.29}  & \multicolumn{1}{c|}{\underline{58.16}} & 83.26       & 58.93      \\
\multicolumn{1}{l|}{Qwen3-1.7B}              & 68.43  & \multicolumn{1}{c|}{39.26} & 68.56  & \multicolumn{1}{c|}{37.48} & 75.58  & \multicolumn{1}{c|}{35.86} & 65.51  & \multicolumn{1}{c|}{30.86} & 67.12  & \multicolumn{1}{c|}{31.18} & 69.04       & 34.93      \\
\multicolumn{1}{l|}{Qwen3-4B}                & 79.99  & \multicolumn{1}{c|}{48.30} & 75.47  & \multicolumn{1}{c|}{46.37} & 80.24  & \multicolumn{1}{c|}{41.84} & 74.04  & \multicolumn{1}{c|}{40.39} & 74.55  & \multicolumn{1}{c|}{40.06} & 76.86       & 43.39      \\
\multicolumn{1}{l|}{Qwen3-4B-Ins}            & 81.96  & \multicolumn{1}{c|}{54.12} & 77.49  & \multicolumn{1}{c|}{52.02} & 82.64  & \multicolumn{1}{c|}{46.37} & 78.03  & \multicolumn{1}{c|}{44.91} & 77.28  & \multicolumn{1}{c|}{49.11} & 79.48       & 49.31      \\
\multicolumn{1}{l|}{Qwen3-8B}                & 81.35  & \multicolumn{1}{c|}{57.99} & 78.43  & \multicolumn{1}{c|}{52.83} & 83.97  & \multicolumn{1}{c|}{49.60} & 79.32  & \multicolumn{1}{c|}{48.47} & 78.08  & \multicolumn{1}{c|}{50.88} & 80.23       & 51.95      \\
\multicolumn{1}{l|}{Qwen3-32B}               & 83.90  & \multicolumn{1}{c|}{60.09} & 79.86  & \multicolumn{1}{c|}{58.64} & 85.15  & \multicolumn{1}{c|}{55.57} & 82.02  & \multicolumn{1}{c|}{53.96} & 80.34  & \multicolumn{1}{c|}{51.86} & 82.25       & 56.02      \\
\multicolumn{1}{l|}{Qwen3-30B-A3B}           & 83.41  & \multicolumn{1}{c|}{58.48} & 79.27  & \multicolumn{1}{c|}{58.16} & 83.75  & \multicolumn{1}{c|}{54.93} & 81.25  & \multicolumn{1}{c|}{53.80} & 79.54  & \multicolumn{1}{c|}{51.70} & 81.44       & 55.41      \\
\multicolumn{1}{l|}{Qwen3-30B-A3B-Ins}       & 84.25  & \multicolumn{1}{c|}{60.90} & 80.49  & \multicolumn{1}{c|}{58.18} & 85.76  & \multicolumn{1}{c|}{54.28} & 82.16  & \multicolumn{1}{c|}{58.00} & 80.50  & \multicolumn{1}{c|}{55.90} & 82.63       & 57.45      \\
\multicolumn{1}{l|}{Qwen3-235B-A22B}         & \textbf{85.33}  & \multicolumn{1}{c|}{\underline{65.43}} & \textbf{82.39}  & \multicolumn{1}{c|}{\underline{64.30}} & \underline{86.57}  & \multicolumn{1}{c|}{\underline{60.26}} & \textbf{84.63}  & \multicolumn{1}{c|}{\underline{60.26}} & 81.59  & \multicolumn{1}{c|}{54.93} & \textbf{84.10}       & \underline{61.04}      \\ \bottomrule
\end{tabular}}
\caption{Results for Fun Posts. SSR denotes Slang Success Rate. Within each model category, the best and second-best scores are highlighted in \textbf{bold} and \underline{underlined} font, respectively.}
\label{funpostsresult}
\end{table*}
\subsection{Experimental Settings}
\subsubsection{Evaluated Models}
We evaluate a broad set of models, covering both traditional MT architectures and modern LLMs. 
The full list of 22 evaluated models is as follows:
\begin{itemize}
    \item \textbf{Closed-source LLM APIs}: GPT-4o~\citep{openai2024gpt4ocard}, GPT-5~\citep{openai2025gpt5}, Claude-Sonnet-4~\citep{Claude4};
    \item \textbf{Open-source general-purpose LLMs}: Deepseek-V3~\citep{deepseekai2025deepseekv3technicalreport}, GPT-OSS-120B~\citep{openai2025gptoss120bgptoss20bmodel}, Aya-101~\citep{2024aya101modelinstructionfinetuned}, Aya-Expanse-8B~\citep{dang2024ayaexpansecombiningresearch}, three Gemma3-IT~\citep{gemmateam2025gemma3technicalreport} series models (4B, 12B, 27B), and eight Qwen3~\citep{yang2025qwen3technicalreport} series models (1.7B, 4B, 4B-Instruct, 8B, 32B, 30B-A3B, 30B-A3B-Instruct, 235B-A22B);
    \item \textbf{Translation-specialized Models}: NLLB-3.3B~\citep{nllbteam2022languageleftbehindscaling}, Hunyuan-MT-7B~\citep{zheng2025hunyuanmttechnicalreport}, GemmaX2-9B~\citep{gemmax2cui2025multilingualmachinetranslationopen} and Google Translate\footnote{\url{https://docs.cloud.google.com/translate/docs/reference/rpc/google.cloud.translate.v2}}.
\end{itemize}

This selection allows us to examine model performance across a range of architectures and model scales, providing a comprehensive view of MT capabilities on Chinese social media text.
\subsubsection{Implementation Details}
We use XCOMET-XL\footnote{\url{https://huggingface.co/Unbabel/XCOMET-XL}} for overall translation quality evaluation. For Fun Posts, the threshold for fuzzy matching is set to 0.8 to tolerate surface variations that should not be penalized (tense, pluralization, abbreviation, details in Appendix~\ref{fuzzy_matching_0.8}). For Social Snippets, style embeddings are obtained using mStyleDistance~\citep{qiu-etal-2025-mstyledistance}\footnote{\url{https://huggingface.co/StyleDistance/mstyledistance}}, emotion embeddings are produced using the model from~\citet{bianchi2021feelxlm-elo}\footnote{\url{https://huggingface.co/MilaNLProc/xlm-emo-t}}
, and sentiment embeddings are produced using \citet{tabularisai_2025-sentimentembed}\footnote{\url{https://huggingface.co/tabularisai/multilingual-sentiment-analysis}}. 
Other implementation details can be found in Appendix~\ref{a2}.

\subsection{Main Experimental Results}
\subsubsection{Fun Posts}

We summarize the results for Fun Posts in Table~\ref{funpostsresult}. For clarity, we divide the table into: (1) closed-source LLM APIs, (2) translation-specialized models, and (3) open-source general-purpose LLMs.

\begin{table*}[t]
\centering
\resizebox{0.99\textwidth}{!}{
\begin{tabular}{lccccccccccccc}
\toprule
\multicolumn{1}{c|}{\multirow{2}{*}{Models}} & \multicolumn{2}{c|}{$zh \to es$}          & \multicolumn{2}{c|}{$zh \to fr$}          & \multicolumn{2}{c|}{$zh \to ja$}          & \multicolumn{2}{c|}{$zh \to ko$}          & \multicolumn{2}{c|}{$zh \to ru$}          & \multicolumn{3}{c}{avg.}                   \\ \cmidrule(lr){2-3} \cmidrule(lr){4-5} \cmidrule(lr){6-7} \cmidrule(lr){8-9} \cmidrule(lr){10-11} \cmidrule(lr){12-14}
\multicolumn{1}{c|}{}                        & XCOMET & \multicolumn{1}{c|}{ES}    & XCOMET & \multicolumn{1}{c|}{ES}    & XCOMET & \multicolumn{1}{c|}{ES}    & XCOMET & \multicolumn{1}{c|}{ES}    & XCOMET & \multicolumn{1}{c|}{ES}    & XCOMET & ES    & \multicolumn{1}{l}{GEMBA} \\ \midrule
\multicolumn{14}{c}{\textit{Closed-source LLM APIs}}                                                                                                                                                                                                                                                 \\ \midrule
\multicolumn{1}{l|}{GPT-4o}                   & \underline{76.67}  & \multicolumn{1}{c|}{\underline{67.84}} & \underline{71.60}  & \multicolumn{1}{c|}{\textbf{67.82}} & \textbf{77.83}  & \multicolumn{1}{c|}{67.27} & 81.77  & \multicolumn{1}{c|}{\textbf{67.25}} & \underline{74.61}  & \multicolumn{1}{c|}{\textbf{70.83}} & \underline{76.50}  & \underline{68.20} & \underline{3.65}                      \\
\multicolumn{1}{l|}{GPT-5}                    & \textbf{77.18}  & \multicolumn{1}{c|}{67.54} & \textbf{71.91}  & \multicolumn{1}{c|}{\underline{67.80}} & \underline{77.21}  & \multicolumn{1}{c|}{\textbf{68.16}} & \underline{81.81}  & \multicolumn{1}{c|}{\underline{67.24}} & \textbf{74.63}  & \multicolumn{1}{c|}{\underline{70.63}} & \textbf{76.55}  & \textbf{68.27} & \textbf{3.69}                      \\
\multicolumn{1}{l|}{Claude-Sonnet-4}         & 75.04  & \multicolumn{1}{c|}{\textbf{68.83}} & 68.37  & \multicolumn{1}{c|}{67.47} & 76.81  & \multicolumn{1}{c|}{\underline{67.74}} & \textbf{81.98}  & \multicolumn{1}{c|}{66.35} & 73.48  & \multicolumn{1}{c|}{70.12} & 75.14  & 68.10 & 3.58                      \\ \midrule
\multicolumn{14}{c}{\textit{Translation-specialized Models}}                                                                                                                                                                                                                                         \\ \midrule
\multicolumn{1}{l|}{NLLB-3.3B}               & 64.00  & \multicolumn{1}{c|}{63.77} & 54.49  & \multicolumn{1}{c|}{61.55} & 53.94  & \multicolumn{1}{c|}{59.55} & 59.01  & \multicolumn{1}{c|}{59.80} & 60.39  & \multicolumn{1}{c|}{64.03} & 58.37  & 61.74 & 2.13                      \\
\multicolumn{1}{l|}{google-translate}        & 75.68  & \multicolumn{1}{c|}{66.51} & \textbf{71.54}  & \multicolumn{1}{c|}{66.04} & 71.67  & \multicolumn{1}{c|}{64.62} & 77.64  & \multicolumn{1}{c|}{64.50} & \underline{72.51}  & \multicolumn{1}{c|}{\underline{68.39}} & 73.81  & 66.01 & 3.17                      \\
\multicolumn{1}{l|}{GemmaX2-9B}                 & \underline{76.64}  & \multicolumn{1}{c|}{\underline{67.13}}     & 70.26  & \multicolumn{1}{c|}{\underline{66.69}}     & \underline{75.98}  & \multicolumn{1}{c|}{\underline{65.34}}     & \underline{79.84}  & \multicolumn{1}{c|}{\underline{65.98}}     & 72.00  & \multicolumn{1}{c|}{68.07}     & \underline{74.94}  & \underline{66.64}     & \underline{3.19}                      \\
\multicolumn{1}{l|}{Hunyuan-MT-7B}           & \textbf{77.10}  & \multicolumn{1}{c|}{\textbf{67.21}} & \underline{70.59}  & \multicolumn{1}{c|}{\textbf{66.84}} & \textbf{76.62}  & \multicolumn{1}{c|}{\textbf{65.35}} & \textbf{81.49}  & \multicolumn{1}{c|}{\textbf{66.04}} & \textbf{73.86}  & \multicolumn{1}{c|}{\textbf{68.79}} & \textbf{75.93}  & \textbf{66.85} & \textbf{3.28}                      \\ \midrule
\multicolumn{14}{c}{\textit{Open-source General-purpose LLMs}}                                                                                                                                                                                                                                             \\ \midrule
\multicolumn{1}{l|}{DeepSeek-V3}             & \textbf{77.32}  & \multicolumn{1}{c|}{\textbf{68.31}} & \textbf{72.92}  & \multicolumn{1}{c|}{\textbf{67.51}} & \textbf{78.30}  & \multicolumn{1}{c|}{67.45} & \underline{81.66}  & \multicolumn{1}{c|}{\underline{66.97}} & \textbf{74.70}  & \multicolumn{1}{c|}{70.17} & \textbf{76.98}  & \textbf{68.08} & \textbf{3.64}                      \\
\multicolumn{1}{l|}{GPT-OSS-120B}            & \underline{77.03}  & \multicolumn{1}{c|}{66.89} & \underline{72.00}  & \multicolumn{1}{c|}{66.97} & \underline{76.93}  & \multicolumn{1}{c|}{65.57} & \textbf{81.69}  & \multicolumn{1}{c|}{66.59} & \underline{73.68}  & \multicolumn{1}{c|}{69.21} & \underline{76.27}  & 67.05 & \underline{3.57}                      \\
\multicolumn{1}{l|}{Aya-101}                 & 69.38  & \multicolumn{1}{c|}{61.20} & 62.73  & \multicolumn{1}{c|}{63.17} & 69.39  & \multicolumn{1}{c|}{60.51} & 72.54  & \multicolumn{1}{c|}{63.05} & 67.52  & \multicolumn{1}{c|}{64.79} & 68.31  & 62.54 & 2.28                      \\
\multicolumn{1}{l|}{Aya-Expanse-8B}          & 72.33  & \multicolumn{1}{c|}{66.23}     & 66.42  & \multicolumn{1}{c|}{65.01}     & 71.19  & \multicolumn{1}{c|}{64.99}     & 76.99  & \multicolumn{1}{c|}{65.03}     & 68.65  & \multicolumn{1}{c|}{68.05}     & 71.12  & 65.86     & 3.14                      \\
\multicolumn{1}{l|}{Gemma3-4B}               & 71.35  & \multicolumn{1}{c|}{65.19} & 64.88  & \multicolumn{1}{c|}{64.39} & 68.85  & \multicolumn{1}{c|}{64.86} & 70.98  & \multicolumn{1}{c|}{64.33} & 66.79  & \multicolumn{1}{c|}{67.21} & 68.57  & 65.20 & 2.85                      \\
\multicolumn{1}{l|}{Gemma3-12B}              & 74.77  & \multicolumn{1}{c|}{65.90} & 68.57  & \multicolumn{1}{c|}{65.58} & 73.31  & \multicolumn{1}{c|}{65.30} & 77.99  & \multicolumn{1}{c|}{65.21} & 70.47  & \multicolumn{1}{c|}{67.37} & 73.02  & 65.87 & 3.20                      \\
\multicolumn{1}{l|}{Gemma3-27B}              & 74.86  & \multicolumn{1}{c|}{66.33} & 69.22  & \multicolumn{1}{c|}{65.93} & 74.18  & \multicolumn{1}{c|}{65.75} & 79.08  & \multicolumn{1}{c|}{65.58} & 71.90  & \multicolumn{1}{c|}{67.69} & 73.85  & 66.26 & 3.37                      \\
\multicolumn{1}{l|}{Qwen3-1.7B}              & 68.24  & \multicolumn{1}{c|}{65.65} & 61.38  & \multicolumn{1}{c|}{65.23} & 67.97  & \multicolumn{1}{c|}{65.09} & 64.73  & \multicolumn{1}{c|}{64.13} & 63.65  & \multicolumn{1}{c|}{67.28} & 65.19  & 65.48 & 2.50                      \\
\multicolumn{1}{l|}{Qwen3-4B}                & 72.41  & \multicolumn{1}{c|}{66.01} & 66.68  & \multicolumn{1}{c|}{66.34} & 68.36  & \multicolumn{1}{c|}{66.08} & 71.66  & \multicolumn{1}{c|}{65.21} & 67.47  & \multicolumn{1}{c|}{69.22} & 69.32  & 66.57 & 2.79                      \\
\multicolumn{1}{l|}{Qwen3-4B-Ins}            & 73.38  & \multicolumn{1}{c|}{67.15} & 68.16  & \multicolumn{1}{c|}{67.07} & 72.20  & \multicolumn{1}{c|}{66.83} & 75.32  & \multicolumn{1}{c|}{65.80} & 68.83  & \multicolumn{1}{c|}{69.50} & 71.58  & 67.27 & 3.03                      \\
\multicolumn{1}{l|}{Qwen3-8B}                & 74.06  & \multicolumn{1}{c|}{66.53} & 67.84  & \multicolumn{1}{c|}{66.96} & 72.75  & \multicolumn{1}{c|}{67.10} & 76.09  & \multicolumn{1}{c|}{66.20} & 69.93  & \multicolumn{1}{c|}{69.66} & 72.13  & 67.29 & 3.06                      \\
\multicolumn{1}{l|}{Qwen3-32B}               & 74.35  & \multicolumn{1}{c|}{66.98} & 69.29  & \multicolumn{1}{c|}{66.93} & 73.90  & \multicolumn{1}{c|}{\textbf{67.55}} & 77.67  & \multicolumn{1}{c|}{66.69} & 70.91  & \multicolumn{1}{c|}{\underline{70.23}} & 73.22  & 67.68 & 3.24                      \\
\multicolumn{1}{l|}{Qwen3-30B-A3B}           & 74.14  & \multicolumn{1}{c|}{66.33} & 68.69  & \multicolumn{1}{c|}{66.82} & 73.10  & \multicolumn{1}{c|}{66.81} & 76.87  & \multicolumn{1}{c|}{66.78} & 70.68  & \multicolumn{1}{c|}{69.07} & 72.70  & 67.16 & 3.17                      \\
\multicolumn{1}{l|}{Qwen3-30B-A3B-Ins}       & 75.34  & \multicolumn{1}{c|}{67.27} & 69.56  & \multicolumn{1}{c|}{67.28} & 74.75  & \multicolumn{1}{c|}{66.64} & 78.55  & \multicolumn{1}{c|}{66.88} & 71.25  & \multicolumn{1}{c|}{69.55} & 73.89  & 67.52 & 3.32                      \\
\multicolumn{1}{l|}{Qwen3-235B-A22B}         & 76.92  & \multicolumn{1}{c|}{\underline{67.38}} & 70.94  & \multicolumn{1}{c|}{\underline{67.44}} & 74.82  & \multicolumn{1}{c|}{\underline{67.51}} & 80.80  & \multicolumn{1}{c|}{\textbf{67.23}} & 72.23  & \multicolumn{1}{c|}{\textbf{70.53}} & 75.14  & \underline{68.02} & 3.41                      \\ \bottomrule
\end{tabular}}
\caption{Results for Social Snippets. ES denotes Embedding Similarity. Within each category, the best and second-best scores are highlighted in \textbf{bold} and \underline{underlined} font, respectively. Full GEMBA results are in Table~\ref{gemba_results}.}
\label{socialsnippetresult}
\end{table*}
\paragraph{Closed-source LLM APIs} exhibit strong translation quality across all five directions. On XCOMET, GPT-4o and GPT-5 perform comparably (the difference is not statistically significant between the two models, Table~\ref{tab:significance}), while GPT-5 leads by a large and consistent margin on SSR, suggesting superior handling of slang and neologisms in informal social media text. Claude-Sonnet-4 performs slightly below the OpenAI models on XCOMET but surpasses GPT-4o on SSR, highlighting its strength in capturing colloquial and stylistic aspects of online discourse. This trend aligns with the observations from~\citet{kocmi2025preliminaryrankingwmt25general}. Across languages, models show complementary strengths. For example, Claude-Sonnet-4 performs particularly well on zh$\to$ko, while all APIs demonstrate strong results on zh$\to$ja, reflecting the relatively close language similarity between Chinese and Japanese.

\paragraph{Translation-specialized models} reveal a consistent gap compared with closed-source LLM APIs. Traditional encoder-decoder models such as NLLB perform noticeably worse across most metrics. Among other models, Hunyuan-MT-7B stands out with robust XCOMET performance across all language pairs, achieving quality comparable to larger models. GemmaX2-9B also performs well in both semantic correctness and slang preservation. Google Translate achieves the highest SSR within this group, indicating relatively strong preservation of informal social-media expressions. Nevertheless, translation-specialized systems still lag behind closed-source LLMs in SSR.



\paragraph{Open-source general-purpose LLMs} present a wide performance spectrum, with larger models generally outperforming smaller ones. DeepSeek-V3 leads this group, achieving the highest XCOMET and SSR among open-source models and surpassing GPT-4o in SSR. Qwen3-235B-A22B also ranks among the strongest systems overall. Within the Gemma3 and Qwen3 families, performance improves consistently with model scale, and recent instruction-tuned variants outperform earlier versions. This trend suggests that both scaling and advanced post-training significantly improve translation robustness~\citep{yang2025qwen3technicalreport}.

Statistical significance tests on our results are provided in Appendix~\ref{statistical}.

\subsubsection{Social Snippets}

We summarize the results for Social Snippets, including XCOMET, the combined embedding similarity~(ES) and Gemba-stars~(GEMBA) in Table~\ref{socialsnippetresult}. Overall, XCOMET scores are noticeably lower than those on Fun Posts, highlighting the difficulty of translating short, informal, and emotionally expressive social media texts.


\paragraph{Closed-source LLM APIs} continue to achieve strong performance. Unlike in Fun Posts, GPT-5 surpasses GPT-4o across both XCOMET and ES, suggesting stronger capability in capturing emotional nuance and informal tone. GEMBA evaluations closely follow the ES trend, further confirming GPT-5’s advantage in preserving stylistic aspects of short social-media texts~\citep{openai2025gpt5}.


\paragraph{Translation-specialized Models} show similar patterns to those observed on Fun Posts. Hunyuan-MT-7B achieves the strongest performance among this group across all metrics, with GemmaX2-9B ranking slightly behind. Google Translate preserves the meaning of slang reasonably well but struggles with informal tone and stylistic nuances. Traditional encoder-decoder models such as NLLB-3.3B perform substantially worse on this dataset.

\vspace{-0.1cm}
\paragraph{Open-source General-purpose LLMs} again exhibit clear scaling trends, with larger models consistently outperforming smaller ones. DeepSeek-V3 leads this group across all metrics and achieves performance comparable to closed-source APIs. Other trends observed on Fun Posts, such as the benefits of instruction tuning, remain consistent.

Overall, the combined embeddings provide a reasonable assessment of translations’ style and emotional fidelity, largely consistent with GEMBA-stars evaluations. Considering the computational resources required for using LLMs as evaluators, ES offers an efficient yet effective alternative. More discussion on ES is provided in Appendix~\ref{embedding similarity}, and we provide human judgments on these automatic metrics in Appendix~\ref{human_annotation}.

\subsection{Analysis}
\subsubsection{Prompt Sensitivity Analysis}
\label{prompt_section}
Prior work has shown that prompting plays a critical role in enhancing translation performance, for example by explicitly providing definitions of culture-specific words or phrases~\citep{ghazvininejad2023dictionarybasedphraselevelpromptinglarge, yao-etal-2024-benchmarking-cultural-awareness}. However, for Chinese social-media-specific slang and neologisms, whose meanings are highly implicit and evolving, manually supplying accurate definitions is both difficult and labor-intensive.
\zky{Instead, we investigate prompt sensitivity on \benchname~through five prompting strategies: (1) Direct Instruction (en), (2) Reminder Prompt, (3) Direct Instruction (zh), (4) Two-stage Reasoning and (5) Few-shot. Details for these prompts are provided in Appendix~\ref{a2}. In addition, for Social Snippets, we select 100 samples whose original posts can be retrieved and incorporate the corresponding posts as additional context for translating comments (denoted as w/o or w/ context).
We specifically choose Qwen3-4B-Ins, Qwen3-8B and GPT-4o for these experiments and the results are reported in Table~\ref{prompt}.}

\begin{table}[t]
\centering
\resizebox{0.49\textwidth}{!}{
\begin{tabular}{l|cc|ccc}
\toprule
\multirow{2}{*}{Models} & \multicolumn{2}{c|}{Fun Posts}               & \multicolumn{3}{c}{Social Snippets}                                \\ \cmidrule(lr){2-3} \cmidrule(lr){4-6}
                        & XCOMET               & SSR                   & XCOMET               & ES                   & GEMBA                \\ \midrule
Qwen3-4B-ins      & & & & & \\
\ \ \ + Direct (en)  & 79.48                & 49.31                 & 71.58                & 67.27                & 3.03                 \\
\ \ \ + Reminder                & \cellcolor{lightgreen}79.65                & \cellcolor{midgreen}50.01                 & \cellcolor{lightred}71.35                &   \cellcolor{lightgrey}67.27                   &   \cellcolor{lightgreen}3.04                   \\ 
\ \ \ + Direct (zh)                & \cellcolor{lightred}79.29                & \cellcolor{lightgreen}49.47                 & \cellcolor{lightred}71.50                &   \cellcolor{lightred}67.24                   &   \cellcolor{lightgrey}3.03                   \\ 
\ \ \ + Two-stage                & \cellcolor{midred}77.54                & \cellcolor{midred}48.07                & \cellcolor{lightgreen}71.97                &   \cellcolor{lightgreen}67.31                   &   \cellcolor{lightgreen}3.08                   \\ 
\ \ \ + Few-shot                & \cellcolor{lightred}79.38               & \cellcolor{lightred}49.23                & \cellcolor{midgreen}72.17                &   \cellcolor{lightgreen}67.30                   &   \cellcolor{lightgreen}3.10                   \\ 
\addlinespace
\ \ \ w/o context & - & - & 70.69 & 67.12 & 3.00 \\
\ \ \ w/ context &- &- & \cellcolor{midgreen}72.34 & \cellcolor{lightgreen}67.35 & \cellcolor{lightgreen}3.11 \\

\midrule
Qwen3-8B               & & & & & \\
\ \ \ + Direct (en)  & 80.23                & 51.95                 & 72.13                & 67.29                & 3.06                 \\
\ \ \ + Reminder                & \cellcolor{midgreen}80.93                & \cellcolor{midgreen}53.08                 & \cellcolor{lightgrey}72.13                &   \cellcolor{lightgreen}67.31                   &   \cellcolor{lightgrey}3.06                   \\ 
\ \ \ + Direct (zh)                & \cellcolor{lightgreen}80.29                & \cellcolor{lightred}51.55                 & \cellcolor{lightred}72.07                &   \cellcolor{lightgrey}67.29                   &   \cellcolor{lightred}3.05                   \\ 
\ \ \ + Two-stage                & \cellcolor{midred}78.07                & \cellcolor{midred}51.05                & \cellcolor{midgreen}72.74                &   \cellcolor{lightgreen}67.35                   &   \cellcolor{lightgreen}3.13                   \\ 
\ \ \ + Few-shot                & \cellcolor{lightred}80.15               & \cellcolor{lightgreen}52.17                & \cellcolor{midgreen}73.27                &   \cellcolor{lightgreen}67.40                   &   \cellcolor{lightgreen}3.19                   \\
\addlinespace
\ \ \ w/o context &- &-  & 72.51 & 67.20 & 3.02 \\
\ \ \ w/ context & - & - & \cellcolor{midgreen}73.97 & \cellcolor{lightgreen}67.37 & \cellcolor{lightgreen}3.19 \\
\midrule
GPT-4o         & & & & & \\
\ \ \ + Direct (en) & 85.50                & 64.91                 & 76.50                & 68.20                & 3.65                 \\
\ \ \ + Reminder                & \cellcolor{lightgreen}85.94  & \cellcolor{midgreen}67.08 & \cellcolor{lightgreen}76.65 & \cellcolor{lightgreen}68.23 & \cellcolor{lightgreen}3.72 \\
\ \ \ + Direct (zh)                & \cellcolor{lightred}85.44  & \cellcolor{lightgreen}65.01 & \cellcolor{lightgreen}76.57 & \cellcolor{lightred}68.19 & \cellcolor{lightgrey}3.65 \\
\ \ \ + Two-stage                & \cellcolor{lightred}85.01  & \cellcolor{midred}64.33 & \cellcolor{midgreen}77.12 & \cellcolor{lightgreen}68.33 & \cellcolor{lightgreen}3.77 \\
\ \ \ + Few-shot                & \cellcolor{lightgreen}85.97  & \cellcolor{lightgreen}65.37 & \cellcolor{midgreen}77.24 & \cellcolor{lightgreen}68.35 & \cellcolor{lightgreen}3.77 \\
\addlinespace
\ \ \ w/o context & - & - & 76.40 & 68.21 & 3.66 \\
\ \ \ w/ context & - & - & \cellcolor{midgreen}77.48 & \cellcolor{lightgreen}68.40 & \cellcolor{lightgreen}3.74 \\

\bottomrule
\end{tabular}}
\caption{Average evaluation results under different prompt strategies. Grey indicates unchanged scores; light green and light red denote small changes ($<$0.5), while green and red denote larger changes ($\geq$0.5).}
\label{prompt}
\end{table}
\zky{We observe different prompt sensitivities for Fun Posts and Social Snippets. For Fun Posts, explicitly alerting LLMs to the existence of slang or neologisms (+Reminder) consistently improves both XCOMET and SSR scores, whereas this strategy provides limited benefit for Social Snippets. In contrast, the two-stage reasoning prompts and the few-shot prompts have little or even negative impact on Fun Posts but improve performance on Social Snippets. Finally, switching the prompt language from English to Chinese does not significantly affect the overall performance on either subset. For Social Snippets, incorporating the original post as additional context information can consistently improve the translation quality.}

This analysis characterizes how prompting strategies interact with the two
subsets rather than establishing prompt-robustness. Paired bootstrap tests against the Direct (en) baseline confirm that the largest effects in each direction are significant ($p<0.001$): +Reminder on Fun Posts (Qwen3-8B, GPT-4o) and +Few-shot on Social Snippets (Qwen3-8B, GPT-4o). Differences from +Direct (zh) are negligible in magnitude on both subsets. Prompt design thus does not transfer across subsets in a uniform way.



\subsubsection{Qualitative Analysis}
\zky{To better understand the challenges of translating social media texts and how models behave under different prompting strategies, we present qualitative case studies in this section. We focus on the Qwen3-4B-Ins model for the zh$\to$ja direction. Examples for Fun Posts are shown in Figure~\ref{fig:case_study_funpost1} and Figure~\ref{fig:case_study_funpost2}, Appendix, while examples for Social Snippets are provided in Figure~\ref{fig:case_study_socialsnippets}, Appendix.} 

\zky{The case studies reflect the performance trends observed in Table~\ref{prompt}. As shown in Figure~\ref{fig:case_study_funpost1} and Figure~\ref{fig:case_study_funpost2}, the Reminder prompt helps the model better handle polite or stylistic expressions that commonly appear in formal texts. In contrast, the Two-stage prompting strategy occasionally causes the model to generate non-existent Japanese words or add extra content beyond the source sentence, possibly due to over-reasoning. The Few-shot prompt helps the model handle overused polite expressions in Japanese, but occasionally leads to unnatural word choices.
The Social Snippets example further shows that when the source sentence contains emojis or repeated exclamation marks, even simple expressions can become difficult for models to translate (e.g., the ``田！老！师！我！想！你！了！ (Mr! Tian! I! miss! you!)'' in Figure~\ref{fig:case_study_socialsnippets} can be easily translated if we exclude redundant exclamation marks, but the model fails when given direct instructions). However, this issue can be mitigated through Two-stage reasoning, Few-shot prompting, or by incorporating additional contextual information.}

\subsubsection{Annotation Analysis}
\label{annotation_analysis}
\zky{The annotation of social-media-specific texts is inherently challenging. During the peer-review process, a sample is only finalized once agreement is reached among reviewers. If a reviewer disagrees with the previous translation, the sample is passed to another reviewer for further revision. We measure annotation difficulty using the average number of review rounds required before final agreement. For Fun Posts, the average number of rounds is 1.2, indicating that most translations are accepted after a single peer review. In contrast, Social Snippets require an average of 2.7 rounds before agreement is reached. This substantial difference suggests that Social Snippets contain greater ambiguity and are inherently more challenging for both human annotators and models. In addition, agreement on slang or neologism usage in Fun Posts requires an average of 1.4 rounds. Additional discussions on annotations are provided in the Appendix~\ref{annotation_discussion}.}

\zky{Other analyses, including a discussion of the distinction between the two subsets and the agreement between automatic metrics and human judgments, can be found in Appendix~\ref{distinction} and Appendix~\ref{human_annotation}.}

\section{Conclusion}
In this work, we introduce \benchname, a benchmark for Chinese social-media machine translation, curated with bilingual expert translations and covering complementary scenarios. Fun Posts focus on context-rich, slang- and neologism-heavy content, while Social Snippets emphasize concise, emotion- and style-driven expressions.
Our evaluations reveal differences in how models handle semantic fidelity, stylistic nuances, and social-media-specific language. Our targeted metrics such as Slang Success Rate, embedding-based similarity effectively assess these challenges, highlighting areas where current models excel or fall short. \benchname~thus provides a practical tool for diagnosing and guiding the development of MT systems tailored to real-world Chinese social media.

\section*{Limitations}
While our work provides a comprehensive benchmark for evaluating machine translation on Chinese social media texts, this work has several limitations. First, we only focus on prompting strategies to improve translation quality, such as reminding models of the presence of slang, neologisms, or stylistic requirements. More sophisticated approaches, e.g., slang-aware pretraining, targeted fine-tuning can be investigated in future works. Second, although \benchname~covers five Chinese-foreign language directions, the inclusion of additional directions is constrained by annotation availability and cost. Finally, \benchname~is designed to reveal systematic patterns in how LLMs handle Chinese social-media text, not to produce a definitive system ranking. Several top-tier models score almost identically on some metrics, and we leave the definitive ranking among such closely matched models to future work.
\section*{Ethics Statement}
\benchname~is constructed with careful filtering to mitigate potential ethical concerns, including the removal of sensitive or inappropriate content. All models used in our experiments are evaluated in accordance with their respective licenses. We use ChatGPT only for grammar refinement and language polishing. Our work does not introduce ethical biases but aims to make new, positive contributions to MT community.
\section*{Acknowledgment}
Kaiyan Zhao was supported by JST SPRING, Grant Number JPMJSP2108.
\bibliography{custom}

\appendix

\section{Appendix}

\subsection{Dataset Construction Details}
\label{appendixA1}
\begin{figure}[t]
    \centering  \includegraphics[width=0.49\textwidth]{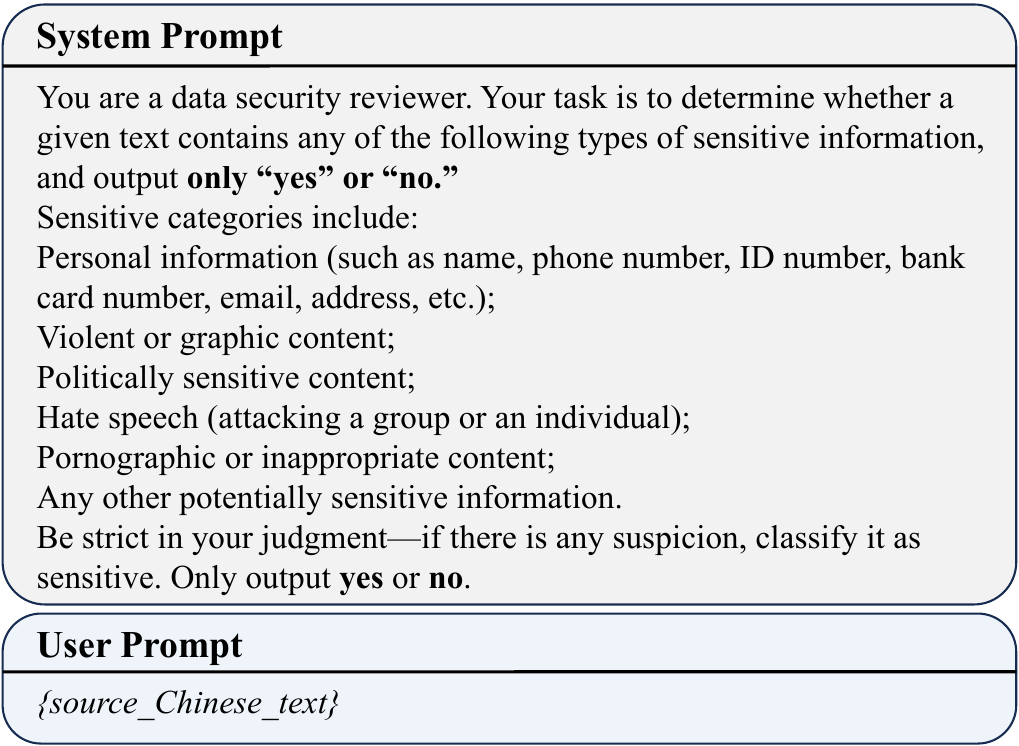}
    \caption{The prompt we used for strict filtering. All samples labeled as ``yes'' are excluded from our dataset. Text enclosed in italic braces \textit{\{…\}} indicates placeholders for the source Chinese content.}
    \label{fig:filter}
\end{figure}

Our candidate data undergo strict filtering to remove: (1) personal information, including names, phone numbers, ID numbers, bank card numbers, email addresses, and physical addresses; (2) violent or graphic content; (3) politically sensitive content; (4) hate speech targeting individuals or groups; and (5) pornographic, inappropriate; or (6) any other potentially sensitive content. The filtering prompt we used is shown in Figure~\ref{fig:filter}. After the LLM-based filtering, all data additionally undergo security screening via internal automated safety APIs to remove any potentially sensitive or risky content.

For Fun Posts, all the 1183 source sentences are shared across all five target languages. For Social Snippets, due to filtering, since short sentences are more difficult to reliably pass quality control, a portion of the source sentences is duplicated across language pairs. \zky{We ensure that all data collection, processing, and usage comply with relevant legal and ethical standards, and no private or personally identifiable information is included in the released dataset. Therefore, the dataset can be safely and legally used for research purposes.}

Following this filtering process, bilingual expert annotators translate the remaining Chinese texts into Spanish, French, Japanese, Korean, and Russian.

\subsection{Annotation}
\label{annotation_discussion}
\zky{Annotators (employees) are assigned with the translation tasks as part of their regular work responsibilities, and no additional payment is provided beyond their standard compensation.
They are proficient in both source and target languages and familiar with Chinese social media expressions, including slang and culture-specific references.} 

\zky{All annotators undergo training before the formal annotation process. We first sample 100 instances for pilot translation and quality checking, during which we refine our translation instructions. This calibration phase also help align annotators’ understanding of how to handle slang, tone preservation, and stylistic differences across subsets.}

\zky{Translation guidelines instruct them to preserve the tone, style, and intent of the original text. Ambiguous expressions or neologisms are carefully interpreted in context to ensure faithful and natural translations. All translations undergo quality control, including peer review and adjudication, to form the final two datasets.} 

\paragraph{Guidelines for human translators:} 
``Translate the Chinese sentence into XXX. Give appropriate translations for potential Chinese slang or neologisms, and preserve the original tone and style as much as possible.'' 

\zky{We intentionally adopt a unified set of translation guidelines for both subsets. The guidelines explicitly emphasize preserving slang expressions, informal tone, and pragmatic nuances whenever possible. The training phase ensures that annotators develop a clear understanding of these requirements.}

\paragraph{Guidelines for peer reviewers:}
``Check whether the translation is accurate and complete with respect to the Chinese source sentence. Revise any incorrect, missing, or inappropriate translations.''

\begin{figure}[t]
    \centering  \includegraphics[width=0.25\textwidth]{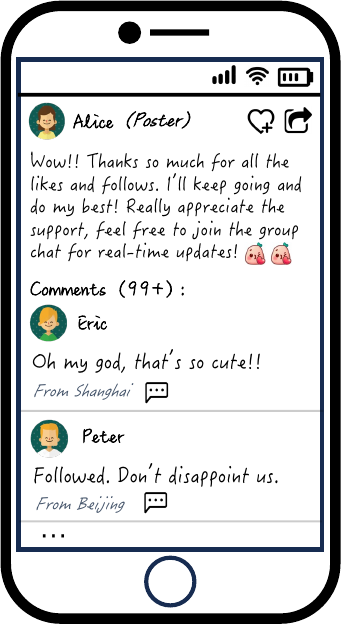}
    \caption{English translation for examples in Figure~\ref{fig:2}.}
    \label{fig:2 translation}
\end{figure}
\begin{table}[ht]
\resizebox{0.49\textwidth}{!}{
\begin{tabular}{lcccccc}
\hline
\multicolumn{1}{c}{Models} & zh$\to$es & zh$\to$fr & zh$\to$ja & zh$\to$ko & zh$\to$ru & avg. \\ \hline
\multicolumn{7}{c}{\textit{Closed-source LLM APIs}}                                \\ \hline
GPT-4o                     & 3.50  & 3.54  & 3.68  & 3.87  & 3.65  & 3.65 \\
GPT-5                      & 3.56  & 3.64  & 3.73  & 3.84  & 3.70  & 3.69 \\
Claude-Sonnet-4            & 3.46  & 3.38  & 3.58  & 3.84  & 3.66  & 3.58 \\ \hline
\multicolumn{7}{c}{\textit{Translation-specialized Models}}                        \\ \hline
NLLB-3.3B                  & 2.19  & 2.14  & 1.97  & 2.17  & 2.18  & 2.13 \\
google-translate           & 3.17  & 3.20  & 3.00  & 3.23  & 3.27  & 3.17 \\
GemmaX2-9B                 & 3.12  & 3.11  & 3.27  & 3.31  & 3.15  & 3.19 \\
Hunyuan-MT-7B              & 3.20  & 3.21  & 3.33  & 3.45  & 3.23  & 3.28 \\ \hline
\multicolumn{7}{c}{\textit{Open-source General-purpose LLMs}}                      \\ \hline
DeepSeek-V3                & 3.58  & 3.55  & 3.65  & 3.79  & 3.63  & 3.64 \\
GPT-OSS-120B               & 3.46  & 3.52  & 3.51  & 3.74  & 3.60  & 3.57 \\
Aya-101                    & 2.20  & 2.21  & 2.53  & 2.29  & 2.18  & 2.28 \\
Aya-Expanse-8B             & 3.04  & 3.12  & 3.14  & 3.27  & 3.13  & 3.14 \\
Gemma3-4B                  & 2.85  & 2.82  & 2.88  & 2.92  & 2.78  & 2.85 \\
Gemma3-12B                 & 3.16  & 3.11  & 3.24  & 3.38  & 3.13  & 3.20 \\
Gemma3-27B                 & 3.26  & 3.31  & 3.38  & 3.54  & 3.37  & 3.37 \\
Qwen3-1.7B                 & 2.46  & 2.56  & 2.71  & 2.36  & 2.41  & 2.50 \\
Qwen3-4B                   & 2.85  & 2.83  & 2.76  & 2.80  & 2.73  & 2.79 \\
Qwen3-4B-Ins               & 3.05  & 3.02  & 3.05  & 3.08  & 2.93  & 3.03 \\
Qwen3-8B                   & 3.05  & 3.01  & 3.10  & 3.14  & 3.00  & 3.06 \\
Qwen3-32B                  & 3.21  & 3.23  & 3.24  & 3.36  & 3.14  & 3.24 \\
Qwen3-30B-A3B              & 3.13  & 3.11  & 3.20  & 3.28  & 3.12  & 3.17 \\
Qwen3-30B-A3B-Ins          & 3.28  & 3.28  & 3.37  & 3.44  & 3.24  & 3.32 \\
Qwen3-235B-A22B            & 3.32  & 3.34  & 3.44  & 3.59  & 3.35  & 3.41 \\ \hline
\end{tabular}}
\caption{Full Gemba-stars results for Social Snippets.}
\label{gemba_results}
\end{table}

\subsection{\zky{Distinction of Fun Posts and Social Snippets}}
\label{distinction}
Beyond text length, different belongings~(posts and comments), and the distribution difference introduced in Section~\ref{dataset_construction}, we separate \benchname~into two subsets for the following reasons.

\subsubsection{Different Roles of Slang}
From a linguistic perspective, slang expressions play different roles in the two subsets. In Fun Posts, slang and neologisms typically appear as identifiable lexical items embedded in longer narrative descriptions. This makes explicit slang identification a meaningful evaluation task.

In contrast, Social Snippets are short reaction-style comments where meaning is often conveyed through overall tone, emotion, and pragmatic style rather than isolated lexical slang units. In such contexts, the stylistic and emotional alignment between source and translation becomes more important than detecting individual slang tokens. Therefore, we evaluate Social Snippets using the ES score, which measures style and emotional consistency between source and target texts.

This distinction is further reflected in the Section~\ref{difficulty}.

\subsubsection{Different Difficulty in Slang and Identification}
\label{difficulty}
We also observe different levels of difficulty in identifying slang and neologisms across the two subsets. We try to identify the slang and neologisms for social snippets, but GPT-5 frequently label the entire sentences as slang or neologisms (in over 50\% of sampled cases), suggesting higher ambiguity in short comment-style texts.

As explained in Section~\ref{annotation_analysis}, we perform human agreement for annotation. During pilot slang identification annotation, we find that the agreement on slang or neologism in Social Snippets often required multiple rounds of discussion (average 3.1 rounds), whereas Fun Posts required significantly fewer revisions (average 1.4 rounds). There even exist cases where annotators cannot make agreements ($<$5\%). This indicates that short comments introduce greater ambiguity for both humans and LLMs in determining what constitutes slang or neologism usage and the reason why we don't apply slang identification on Social Snippets. Thus, separating the subsets therefore helps capture differences in annotation difficulty.

\subsubsection{Unsuitability for ES on Fun Posts}
We also examined the behavior of ES score on the two subsets. While ES reliably distinguishes human and machine translations in Social Snippets, the metric becomes unstable for longer Fun Posts.

One explanation is that long posts often contain multiple emotional tones or discourse intentions within a single sentence. For example, the Chinese sentence in Fun Posts: ``这么鲜活又真实的小作文属实难得一见，像是把日记分享给我们看 希望宝宝越来越好，能不受情绪左右，勇敢做自己！她的每一套穿搭，用的所有小东西都超级无敌种草！'' (\textit{It’s honestly rare to see such a vivid and genuine little post, it feels like she’s sharing her diary with us . I really hope she keeps getting better, stays strong, and can bravely be herself without being swayed by emotions. Every outfit she wears and every little thing she uses makes me want to buy it so badly!}) contains multiple and complex emotions. In such cases, embedding-based classifiers tend to collapse these nuanced signals into a single category, which introduces noise in ES measurements. Due to this instability, we find that ES is more suitable for short, focused comment-style text, and therefore restrict its use to Social Snippets.

\subsubsection{Different Prompt Sensitivity}
Finally, the two subsets exhibit different prompt sensitivities as introduced in Section~\ref{prompt_section}. This difference further supports our decision to evaluate the subsets separately. If we merge two subsets together, we may not observe the different sensitivities.

\subsection{Implementation Details}
\label{a2}
We utilize different prompts depending on model family. For NLLB and Google Translate, we explicitly specify the source and target languages for translation. For Aya-101, we apply a simple translation prompt: 

``Translate to XXX: \textit{\{source text\}}''. 

For the remaining LLMs, we use a unified instruction: 

``You are a translation expert. Please translate the following sentence into XXX and output only the translated result: \textit{\{source text\}}''. 

For models equipped with ``thinking'' modes (e.g., Qwen3), we disable these features to ensure fair comparison. The maximum number of newly generated tokens is set to 512 across all models. We set temperature to 0.0 for API models and use greedy decoding~(\texttt{do\_sample=False}) for other LLMs to ensure reproducibility.
We use GPT-4o for GEMBA-stars evaluation. Filtering is performed with GPT-5. All experiments are conducted on eight H800 GPUs with a single run.

In Section~\ref{prompt_section}, the five prompting strategies we used are as follows:
\begin{itemize}
    \item Direct Instruction (en): Exactly the same English prompt as we use in the main experiments.
    \item Reminder Prompt: For Fun Posts, we use ``You are a translation expert. Please translate the following sentence into XXX. \textbf{Some sentences contain Chinese social media neologisms or slang; when translating, use appropriate expressions.} Output only the translated result: \textit{\{source text\}}''. For Social Snippets, we use ``You are a translation expert. Please translate the following sentence into XXX, \textbf{preserving the tone and style as much as possible.} Output only the translated result: \textit{\{source text\}}''.
    \item Direct Instruction (zh): The Chinese translation for the English direct instruction.
    \item Two-stage Reasoning: For Fun Posts, we use "You are a translation expert. \textbf{First, identify any Chinese internet slang in the following sentence and provide an appropriate XXX expression for it. Then, translate the full sentence into XXX, making sure to preserve suitable expressions for the Chinese internet slang.} Only output the translated result.". For Social Snippets, we use "You are a translation expert. \textbf{First, analyze the theme, tone, and style of the following sentence, then translate it into XXX.} Only output the translation result.".
    \item Few-shot Prompts: For Fun Posts, we use the following 3-shot examples: ``1. 姐妹们，这个面霜真的有被惊艳到！质地很轻薄，上脸一点都不黏，第二天起床皮肤软软的，感觉状态都变好了。2. 第一次自己做蛋糕，没想到居然成功了！虽然卖相一般，但味道真的不错，成就感拉满。 3.本来只是随便试试，结果一试就停不下来，这个真的会上瘾。 and their corresponding translations in five languages''. For Social Snippets, we use "1. 狠狠共鸣了。2. 这也太真实了吧。3. 看完有点被治愈到，谢谢你分享。and their corresponding translations in five languages".
\end{itemize}

For Qwen3 series models, the suffix ``-ins'' is used interchangeably with ``Instruct-2507'', for example, Qwen3-4B-ins equals to Qwen3-4B-Instruct-2507\footnote{\url{https://huggingface.co/Qwen/Qwen3-4B-Instruct-2507}}.
\begin{figure}[t]
    \centering  \includegraphics[width=0.49\textwidth]{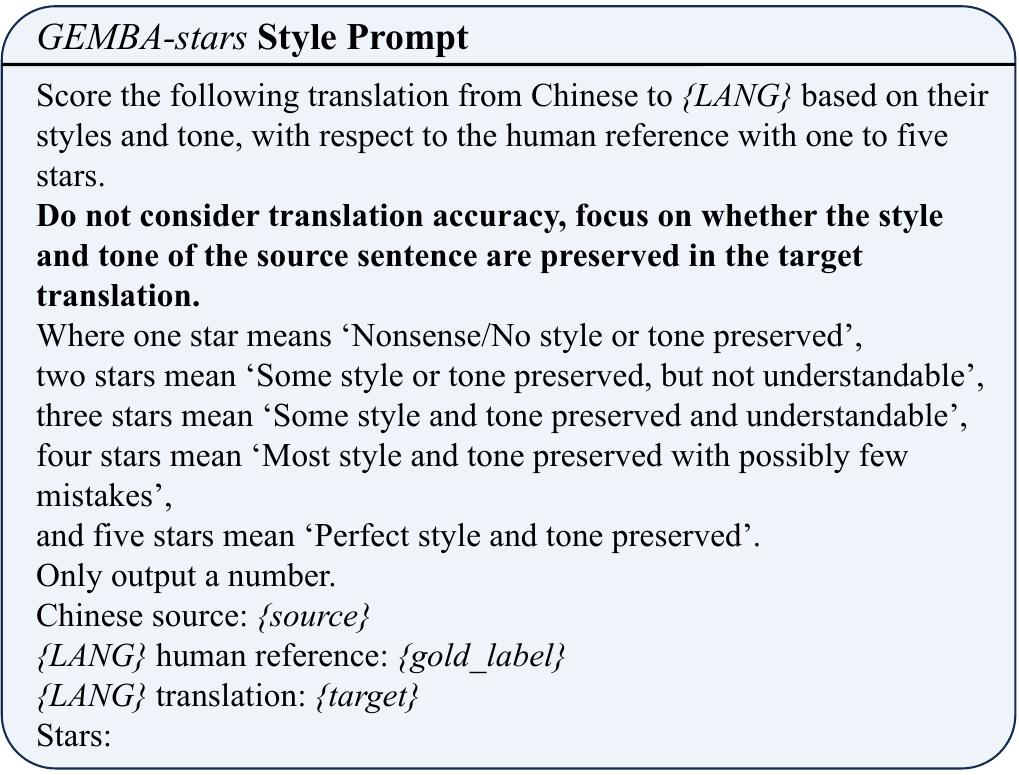}
    \caption{Our GEMBA-stars style prompt. \textbf{Bold} font highlights the difference from the original GEMBA-stars usage.}
    \label{fig:gemba}
\end{figure}

\begin{figure*}[t]
    \centering  \includegraphics[width=0.8\textwidth]{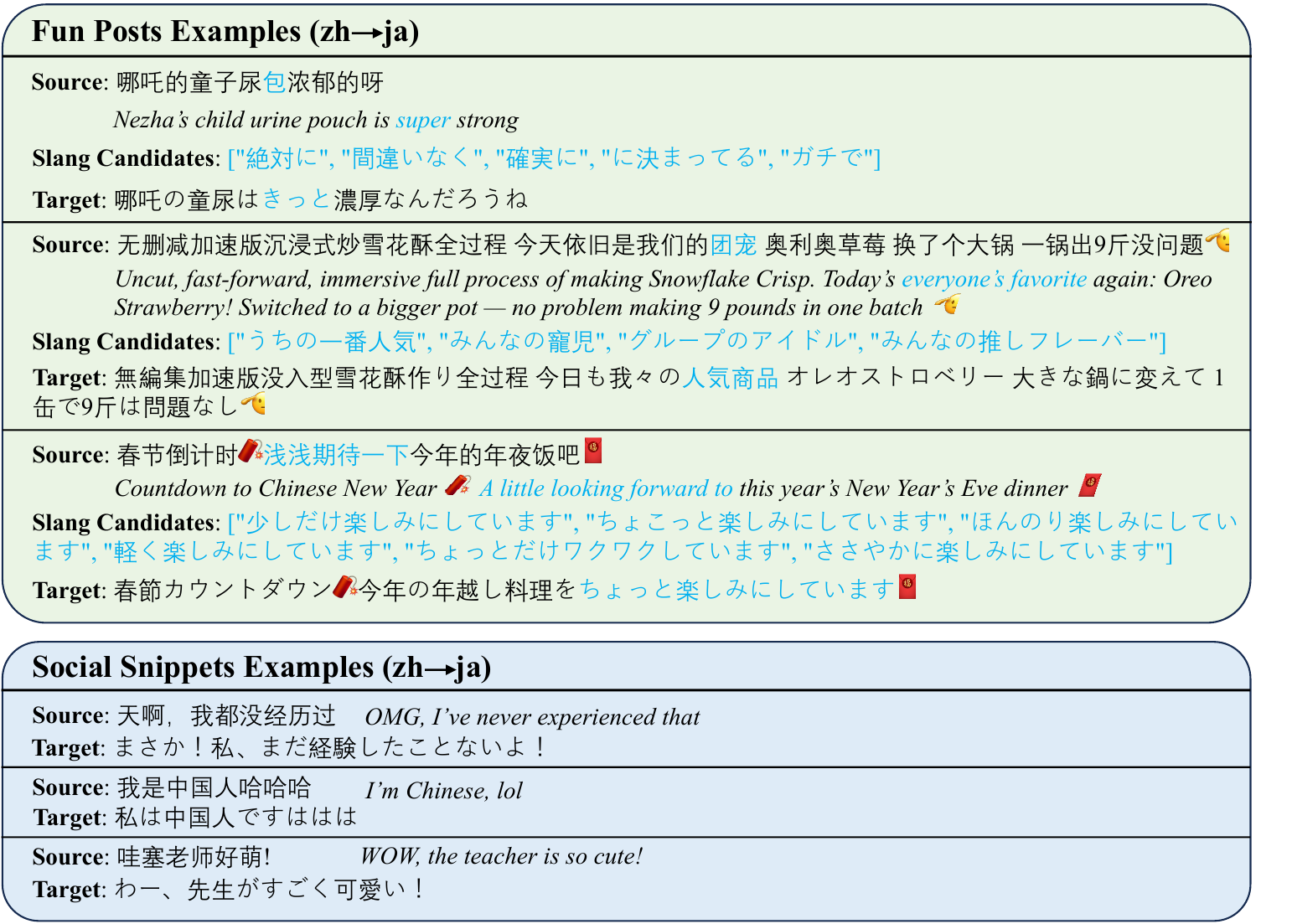}
    \caption{Examples from Fun Posts and Social Snippets (zh→ja). In Fun Posts, we highlight slang and neologisms in \textcolor{cyan}{blue}. \textbf{Target} denotes the human-provided translation. \textbf{Slang Candidates} list alternative slang or neologism choices other than the gold label one in the \textbf{Target}. English references are provided in \textit{italics}. }
    \label{fig:example}
\end{figure*}
\begin{figure}[t]
    \centering  
    \includegraphics[width=0.49\textwidth]{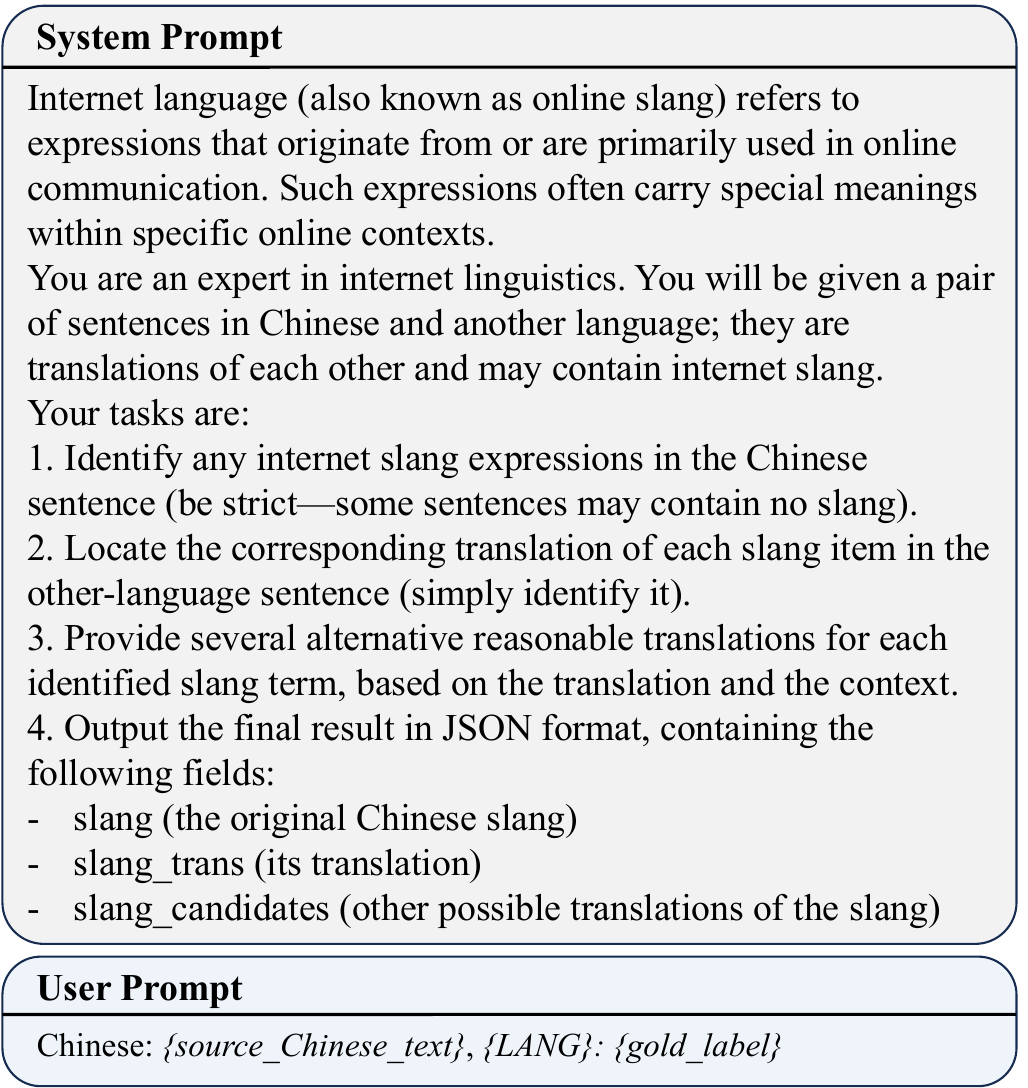}
    \caption{Slang-translation pair detection and slang candidates generation prompt.}
    \label{fig:candidate}
\end{figure}

\subsection{Discussion on Embedding Similarity}
\label{embedding similarity}
In general, embedding-based automatic metrics aim to offer a convenient and computationally efficient alternative to LLM-as-a-judge methods.

The mStyleDistance~\citep{qiu-etal-2025-mstyledistance} we adopt serves as a style embedding model that maps texts with similar stylistic properties to nearby representations while pushing stylistically different texts farther apart, largely independent of semantic content. This design aligns well with our setting, where the goal is to evaluate whether the stylistic characteristics of the source sentence are preserved in the translation~\citep{patel-etal-2025-styleembedding}. 

The emotion embedding model~\citep{bianchi2021feelxlm-elo} classifies sentences into four emotion categories: joy, sadness, anger, and fear. The sentiment embedding model~\citep{tabularisai_2025-sentimentembed} categorizes sentences into five sentiment classes: Neutral, Positive, Negative, Very Negative, and Very Positive. A good translation should preserve the emotion and sentiment of the source sentence.

For these models, we do not directly compare whether the predicted labels of the source and target sentences match, as social media texts often express mixed or nuanced emotions that fall outside the predefined label sets. Instead, we report the cosine similarity between the source and target sentence embeddings to measure the degree of emotional and sentiment alignment.

However, embedding-based automatic metrics still have limitations. As shown in Table~\ref{socialsnippetresult}, the highest ES score is 68.27 for GPT-5, while the lowest score is 61.74 for NLLB-3.3B, indicating a relatively narrow performance gap across models. When computing ES between the source sentences and human-annotated translations, the average ES score reaches 70.32. Although ES exhibits limited absolute discriminability, it remains effective in revealing relative performance differences between models, as its ranking trend closely aligns with GEMBA-stars evaluations. Considering the substantially lower computational cost compared to LLM-as-a-judge methods, we believe ES provides a practical and efficient alternative for large-scale evaluation.

\begin{table*}[t]
\centering
\small
\begin{tabular}{lccccc}
\toprule
\textbf{Comparison} & \textbf{zh$\rightarrow$es} & \textbf{zh$\rightarrow$fr} & \textbf{zh$\rightarrow$ja} & \textbf{zh$\rightarrow$ko} & \textbf{zh$\rightarrow$ru} \\
\midrule
\multicolumn{6}{l}{\textit{Panel A: XCOMET}} \\
\midrule
GPT-4o vs.\ GPT-5              & 0.890 & 0.234 & 0.036 & 0.750 & 0.264 \\
Hunyuan-MT-7B vs.\ GemmaX2-9B  & 0.406 & 0.984 & \textbf{<0.001} & \textbf{<0.001} & \textbf{<0.001} \\
DeepSeek-V3 vs.\ GPT-OSS-120B  & 0.846 & 0.960 & \textbf{<0.001} & 0.222 & \textbf{<0.001} \\
\midrule
\multicolumn{6}{l}{\textit{Panel B: SSR}} \\
\midrule
GPT-4o vs.\ GPT-5              & \textbf{<0.001} & \textbf{<0.001} & \textbf{<0.001} & \textbf{<0.001} & \textbf{<0.001} \\
Hunyuan-MT-7B vs.\ GemmaX2-9B  & 0.984 & 0.352 & 0.306 & 0.132 & 0.934 \\
DeepSeek-V3 vs.\ GPT-OSS-120B  & \textbf{<0.001} & \textbf{<0.001} & 0.010 & \textbf{<0.001} & \textbf{<0.001} \\
\bottomrule
\end{tabular}
\caption{Paired bootstrap $p$-values (10{,}000 resamples, two-sided) for the top-two ranked systems in each model tier on the Fun Posts. \textbf{Bold} indicates significance at $\alpha=0.05$ after Holm correction within each panel (15 tests). Under XCOMET, GPT-4o and GPT-5 are statistically indistinguishable on all five language pairs, whereas SSR separates them on all five.}
\label{tab:significance}
\end{table*}
\subsection{Statistical Significance}
\label{statistical}
We assess the top-two comparison within each model tier using a paired bootstrap test (10,000 resamples at the segment level, two-sided), with Holm correction applied within each metric. Table~\ref{tab:significance} reports the resulting
$p$-values on Fun Posts. Two patterns emerge. First, significance is
\emph{language-dependent}: Hunyuan-MT-7B is ahead of GemmaX2-9B on
zh$\rightarrow$\{ja, ko, ru\} ($p<0.001$) but the two systems are statistically indistinguishable on zh$\rightarrow$\{es, fr\} ($p=0.41$ and $p=0.98$), so an aggregate ranking would obscure a systematic split across target languages.
Second, and more consequentially, significance is \emph{metric-dependent}:
GPT-4o and GPT-5 cannot be separated under XCOMET on any language pair, yet SSR separates them on all five ($p<0.001$). XCOMET measures general adequacy while SSR targets social-media-specific phenomena, and the two therefore differ in
discriminative power precisely on the phenomena CSM-MTBench is designed to
isolate. The reverse case is equally informative: Hunyuan-MT-7B and GemmaX2-9B are cleanly separated by XCOMET on three pairs but by SSR on none, indicating
that dedicated MT models differ in general translation quality without differing in their handling of social-media conventions.

\begin{figure*}[t]
    \centering  
    \includegraphics[width=0.9\textwidth]{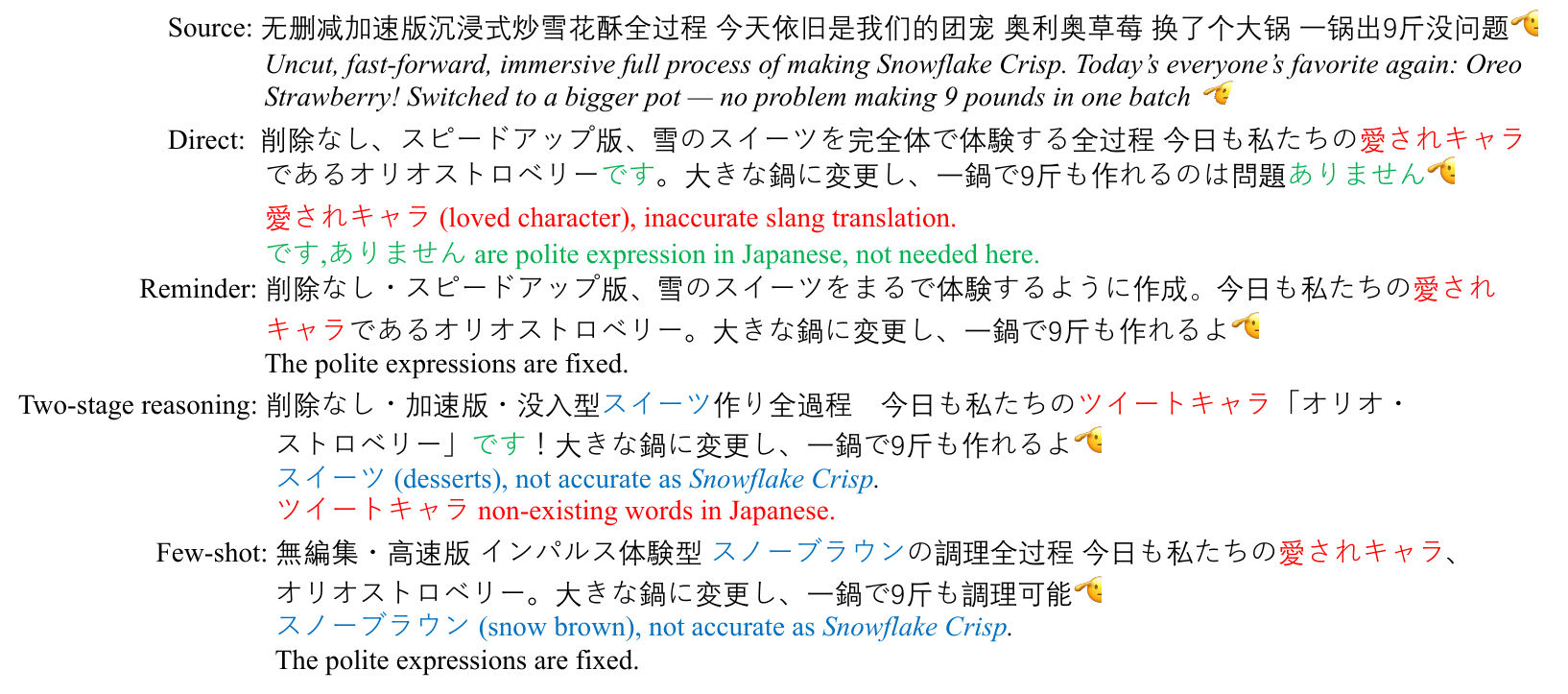}
    \caption{Case study (Part 1) for Fun Posts. Different colors highlight translation issues in the model outputs. Explanations are provided below the Japanese translation.}
    \label{fig:case_study_funpost1}
\end{figure*}
\begin{figure*}[t]
    \centering  
    \includegraphics[width=0.8\textwidth]{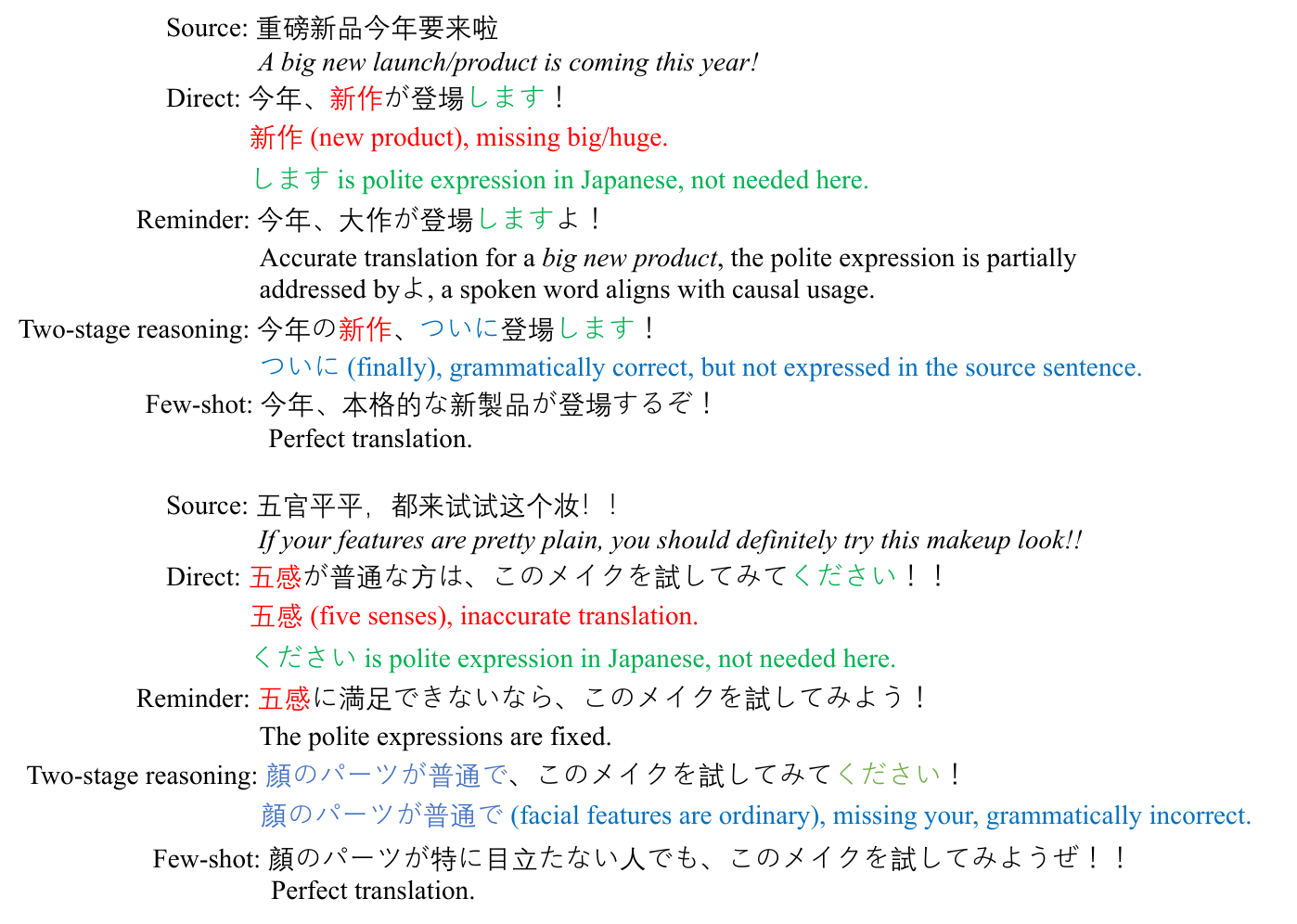}
    \caption{Case study (part 2) for Fun Posts.}
    \label{fig:case_study_funpost2}
\end{figure*}
\begin{figure*}[t]
    \centering  
    \includegraphics[width=0.99\textwidth]{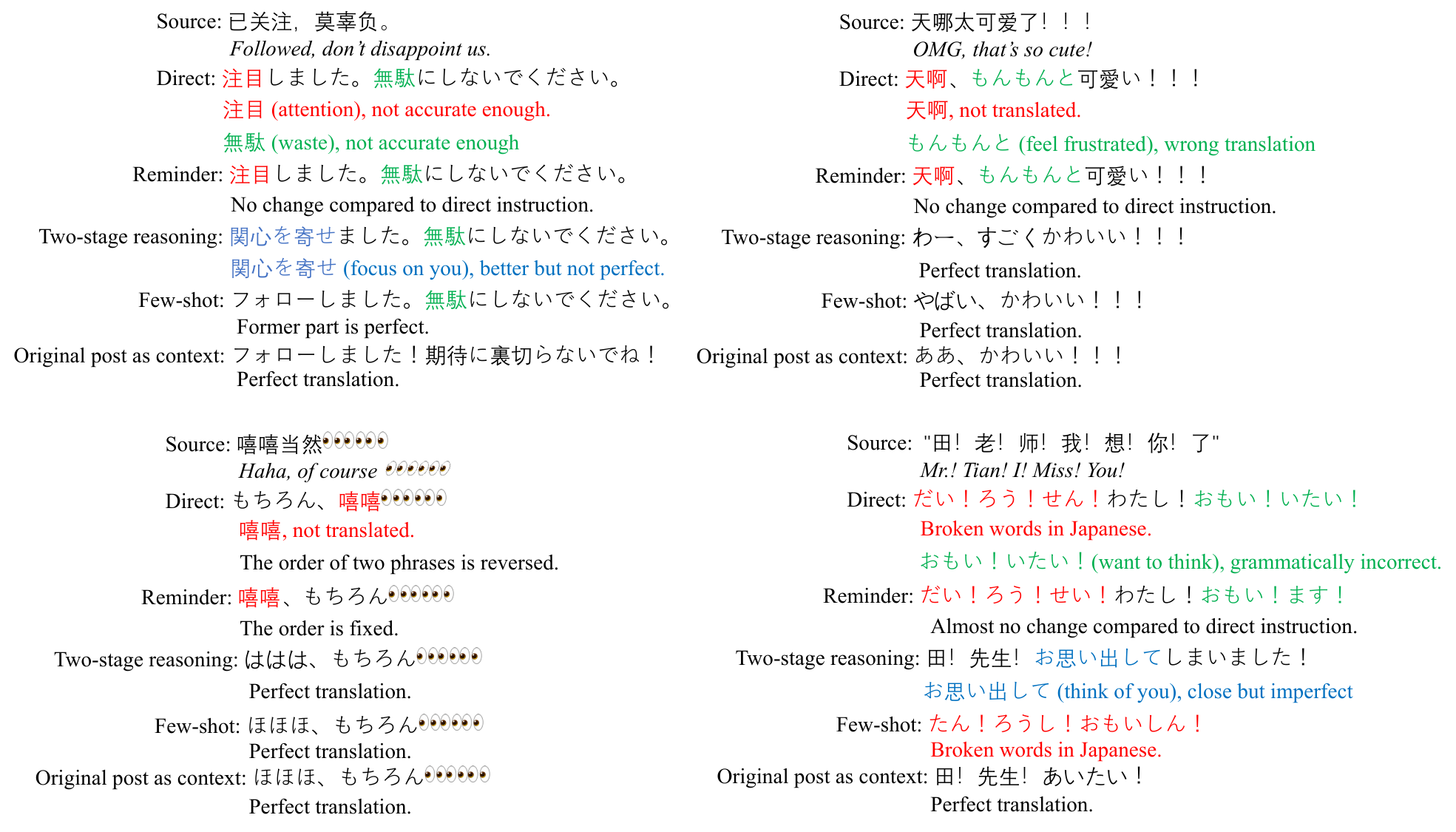}
    \caption{Case study for Social Snippets. Different colors highlight translation issues in the model outputs. Explanations are provided below the Japanese translation.}
    \label{fig:case_study_socialsnippets}
\end{figure*}

\subsection{\zky{Automatic Metrics v.s. Human Judgments}}
\label{human_annotation}
In this section, we compare automatic metrics, including SSR (fuzzy matching), Embedding Similarity (ES) and LLM-as-a-judge to human judgments.
\subsubsection{SSR (fuzzy matching)}
\label{fuzzy_matching_0.8}
The SSR is calculated using fuzzy matching with a threshold of 0.8 (0-1), where scores $\geq$0.80 is treated as a match (score=1) and <0.80 as non-match (score=0). The threshold is chosen to tolerate minor surface variations such as tense differences, pluralization, or small lexical changes, which are common in translation outputs and should not be penalized as incorrect.
For example, the Japanese translation for ``Instagram'' sometimes becomes ``ins'' or ``insta'' in Japanese (both abbreviations are correct), but if the threshold is set to 100 (fuzzy match becomes exact match), the score becomes 0 for this sample.

To assess the reliability of this criterion, we manually inspect 1,000 randomly sampled instances (500 matched and 500 unmatched). We observe a small number of false positives. In several cases, models copy the source sentence instead of translating it, artificially increasing lexical overlap (e.g., leaving “TikTok” untranslated). In other cases, slang expressions are left untranslated, particularly in zh$\to$ja, where shared logographic characters between Chinese and Japanese may make untranslated Chinese slang appear superficially plausible. We also find instances where grammatically incorrect or partially translated outputs were still scored as matches by fuzzy matching.

We also observe a small number of false negatives, where the model produce semantically correct alternatives that fall below the similarity threshold. Some of these cases are due to limitations of the candidate dictionary, which does not cover all valid alternative expressions, while others result from correct translations that differ substantially in surface form from the reference candidates.

Importantly, common fuzzy-matching failure cases such as word-order variation are rarely observed. This is likely because our evaluation focuses on slang expressions and neologisms, which typically function as lexical units rather than compositional phrases, making surface-form matching relatively robust in this setting.

Overall, manual inspection indicates that approximately 4\% of cases show discrepancies between fuzzy matching and human judgment. While imperfect, this relatively low error rate suggests that fuzzy matching provides a reasonably reliable proxy for evaluating slang and neologism translation in our task.

\subsubsection{Embedding Similarity (ES)}
We conducted a human-correlation study to assess the reliability of ES scores. We sample 500 instances in Social Snippets across the five language directions. Five annotators (one per language direction) rated each translation on three dimensions: style, emotion, and sentiment on a 1–5 Likert scale. We report the Spearman correlation to see the rank of ES scores and human judgments, which results in $\rho=0.50, p<0.001$.
For reference, neural metrics on related sentence-level evaluation tasks typically report correlations in a similar range~\citep{freitag-etal-2023-results}. Given that style/emotion preservation is a more subjective task than translation quality assessment, our correlation suggests that ES provides a reasonable proxy for human judgment in our evaluation setting.

\subsubsection{LLM-as-a-judge}
We further assess the reliability of LLM-based evaluation (GEMBA-stars) through manual verification. Specifically, we randomly sample 500 translation instances across the five language directions (100 per direction) and compare the scores assigned by the LLM judge with human judgments. The evaluation uses a 1–5 rating scale following the GEMBA-stars protocol.

Overall, we observe a high level of consistency between the LLM judge and human evaluation. In most cases, the difference between the two ratings is within one point on the 1–5 scale. The average absolute score difference is 0.48, and only a small portion of samples exhibit larger deviations.

We then manually inspect the cases with noticeable discrepancies and find that they typically arise when translations contain subtle stylistic differences or model meets extremely short sentences that are difficult to evaluate consistently. Despite these occasional disagreements, the overall alignment suggests that the LLM-as-a-judge provides a reasonably reliable approximation of human evaluation for our task.

\end{CJK*}
\end{document}